\documentclass[preprint, 12pt]{elsarticle}
\usepackage[T1]{fontenc}    %

\usepackage{fancyhdr}

\usepackage{enumerate}

\usepackage{amsmath,amsthm,amsfonts,amssymb,amscd}
\usepackage{upgreek} %
\usepackage{mathrsfs} %
\usepackage{mathtools} %
\usepackage{derivative}
\makeatletter
\newcommand{\subalign}[1]{%
  \vcenter{%
    \Let@ \restore@math@cr \default@tag
    \baselineskip\fontdimen10 \scriptfont\tw@
    \advance\baselineskip\fontdimen12 \scriptfont\tw@
    \lineskip\thr@@\fontdimen8 \scriptfont\thr@@
    \lineskiplimit\lineskip
    \ialign{\hfil$\m@th\scriptstyle##$&$\m@th\scriptstyle{}##$\hfil\crcr
      #1\crcr
    }%
  }%
}
\makeatother

\usepackage{thmtools,thm-restate}
\makeatletter
\newtheoremstyle{indented}
  {3pt}%
  {3pt}%
  {\addtolength{\@totalleftmargin}{3.5em}
   \addtolength{\linewidth}{-3.5em}
   \parshape 1 3.5em \linewidth}%
  {%
  }%
  {\bfseries}%
  {.}%
  {.5em}%
  {}%
\makeatother

\theoremstyle{plain}
\newtheorem{theorem}{Theorem}[section]

\newtheorem{lemma}[theorem]{Lemma}

\theoremstyle{definition}

\newtheorem{assumption}[theorem]{Assumption}

\theoremstyle{indented}

\theoremstyle{remark}

\usepackage{xcolor}

\usepackage{float}
\usepackage{placeins} %

\usepackage{graphicx}
\usepackage{tikz,tikz-cd}
\usetikzlibrary{arrows, shapes}
\usetikzlibrary{mindmap}
\tikzcdset{scale cd/.style={every label/.append style={scale=#1}, cells={nodes={scale=#1}}}}
\tikzcdset{trapezium stretches=true}
\tikzcdset{arrow style=tikz, diagrams={>=stealth'}}

\usepackage{booktabs}
\usepackage{multirow}
\usepackage{makecell}

\usepackage{caption}
\usepackage{subcaption}

\usepackage[linesnumbered,ruled]{algorithm2e}
\usepackage{listings}

\usepackage[colorlinks=false]{hyperref}

\usepackage{natbib}

\usepackage{lastpage}

\usepackage[capitalise,nameinlink,noabbrev]{cleveref} 
\crefname{section}{Sec.}{Sec.}
\Crefname{section}{Section}{Sections}
\crefname{subsection}{Sec.}{Sec.}
\Crefname{subsection}{Section}{Sections}
\crefname{figure}{Fig.}{Fig.}
\Crefname{figure}{Figure}{Figures}
\crefname{table}{Table}{Tables}
\Crefname{table}{Table}{Tables}
\crefname{equation}{Eq.}{Eq.}
\Crefname{equation}{Equation}{Equations}
\crefname{algocf}{Alg.}{Alg.}
\Crefname{algocf}{Algorithm}{Algorithms}
\crefname{conjecture}{Conjecture}{Conjectures}
\Crefname{conjecture}{Conjecture}{Conjectures}
\crefname{assumption}{Assumption}{Assumptions}
\Crefname{assumption}{Assumption}{Assumptions}
\crefname{objective}{Objective}{Objectives}
\Crefname{objective}{Objective}{Objectives}
\crefname{subobjective}{Objective}{Objectives}
\Crefname{subobjective}{Objective}{Objectives}
\crefname{gap}{Gap}{Gaps}
\Crefname{gap}{Gap}{Gaps}
\crefname{subgap}{Gap}{Gaps}
\Crefname{subgap}{Gap}{Gaps}

\definecolor{UMBlue}{HTML}{00274C}
\definecolor{UMMaize}{HTML}{FFCB05}

\definecolor{UMRed}{HTML}{9A3324}
\definecolor{UMArbBlue}{HTML}{2F65A7}
\definecolor{UMPurple}{HTML}{702082}
\definecolor{UMViolet}{HTML}{575294}
\definecolor{UMStone}{HTML}{655A52}
\definecolor{UMBlack}{HTML}{131516}

\definecolor{UMOrange}{HTML}{D86018}
\definecolor{UMGray}{HTML}{989C97}
\definecolor{UMGreen}{HTML}{75988d}
\definecolor{UMGreenDark}{HTML}{5e7a71}
\definecolor{UMTeal}{HTML}{00B2A9}
\definecolor{UMBeige}{HTML}{9B9A6D}

\definecolor{snsBlue}{HTML}{4c72b0}
\definecolor{snsOrange}{HTML}{dd8452}

\DeclareMathOperator*{\argmax}{arg\,max}
\DeclareMathOperator*{\argmin}{arg\,min}

\renewcommand{\[}{\left[}
\renewcommand{\]}{\right]}
\renewcommand{\(}{\left(}
\renewcommand{\)}{\right)}

\newcommand{\pp}[2]{\frac{\partial #1}{\partial #2}}

\newcommand{\norm}[2]{\left\|\, #1 \,\right\|_{#2}}

\newcommand{\vvvert}{|\kern-1pt|\kern-1pt|}

\newcommand{\DKL}{D_{\mathrm{KL}}}

\newcommand{\bsa}{\boldsymbol{a}}

\newcommand{\bse}{\boldsymbol{e}}

\newcommand{\bsi}{\boldsymbol{i}}
\newcommand{\bsj}{\boldsymbol{j}}

\newcommand{\bsn}{\boldsymbol{n}}

\newcommand{\bsq}{\boldsymbol{q}}
\newcommand{\bsr}{\boldsymbol{r}}
\newcommand{\bss}{\boldsymbol{s}}

\newcommand{\bsw}{\boldsymbol{w}}
\newcommand{\bsx}{\boldsymbol{x}}
\newcommand{\bsy}{\boldsymbol{y}}

\newcommand{\bsA}{\boldsymbol{A}}

\newcommand{\bsI}{\boldsymbol{I}}

\newcommand{\bsT}{\boldsymbol{T}}

\newcommand{\btheta}{\boldsymbol{\theta}}
\newcommand{\bstheta}{\boldsymbol{\theta}}

\newcommand{\bsXi}{\boldsymbol{\Xi}}

\newcommand{\EE}{\mathbb{E}}

\newcommand{\RR}{\mathbb{R}}

\newcommand{\CalA}{\mathcal{A}}

\newcommand{\CalB}{\mathcal{B}}

\newcommand{\CalD}{\mathcal{D}}

\newcommand{\CalM}{\mathcal{M}}

\newcommand{\CalN}{\mathcal{N}}

\newcommand{\CalO}{\mathcal{O}}

\newcommand{\CalQ}{\mathcal{Q}}

\newcommand{\CalR}{\mathcal{R}}

\newcommand{\CalS}{\mathcal{S}}

\newcommand{\CalT}{\mathcal{T}}

\newcommand{\hpsi}{\hat{\psi}}

\journal{Computer Methods in Applied Mechanics and Engineering}
\date{}

\begin{document}

\begin{frontmatter}
\title{FP-IRL: Fokker--Planck Inverse Reinforcement Learning --- A Physics-Constrained Approach to Markov Decision Processes}

\author[UM_ME]{Chengyang Huang}
\author[Auburn]{Siddhartha Srivastava}
\author[UM_R]{Kenneth K. Y. Ho}
\author[UM_R]{Kathryn E. Luker}
\author[UM_R]{Gary D. Luker}
\author[UM_ME]{Xun Huan}
\author[USC]{Krishna Garikipati}

\affiliation[UM_ME]{
    organization={Department of Mechanical Engineering, University of Michigan},
    city={Ann Arbor},
    state={Michigan},
    postcode={48109}, 
    country={United States}
}

\affiliation[Auburn]{
    organization={Department of Aerospace Engineering, Auburn University},
    city={Auburn},
    state={Alabama},
    postcode={36849}, 
    country={United States}
}

\affiliation[UM_R]{
    organization={Department of Radiology, University of Michigan},
    city={Ann Arbor},
    state={Michigan},
    postcode={48109}, 
    country={United States}
}

\affiliation[USC]{
    organization={Department of Aerospace and Mechanical Engineering, University of Southern California},
    city={Los Angeles},
    state={California},
    postcode={90089},
    country={United States}
}

\begin{abstract}
Inverse reinforcement learning (IRL) is a powerful paradigm for uncovering the incentive structure that drives agent behavior, by inferring an unknown reward function from observed trajectories within a Markov decision process (MDP). However, most existing IRL methods require access to the transition function, either prescribed or estimated \textit{a priori}, which poses significant challenges when the underlying dynamics are unknown, unobservable, or not easily sampled.

We propose Fokker--Planck inverse reinforcement learning (FP-IRL), a novel physics-constrained IRL framework tailored for systems that can be described by Fokker--Planck (FP) dynamics. FP-IRL simultaneously infers both the reward and transition functions directly from trajectory data, without requiring access to sampled transitions. Our method leverages a correspondence between MDPs and the FP equation, linking reward maximization in MDPs with free energy minimization in FP dynamics. This connection enables inference of the FP potential function using our inference approach of variational system identification, from which the full set of MDP components---reward, transition, and policy---can be recovered using analytic expressions. 

We demonstrate the effectiveness of FP-IRL through experiments on synthetic benchmarks and a modified version of the Mountain Car problem. Our results show that FP-IRL achieves accurate recovery of agent incentives while preserving computational efficiency and physical interpretability.
\end{abstract}


\begin{graphicalabstract}
    \centering
    \includegraphics[width=\textwidth]{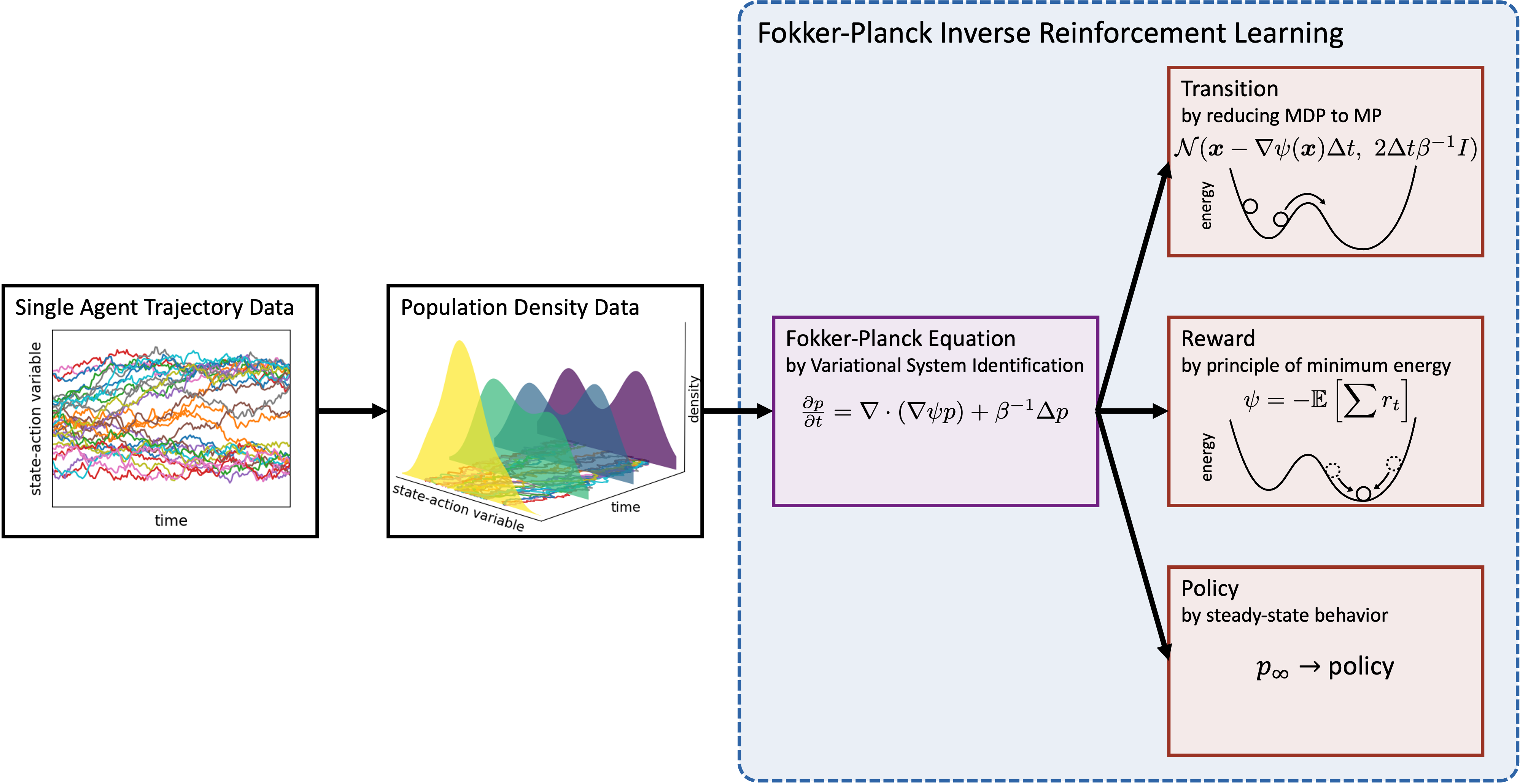}
\end{graphicalabstract}

\begin{highlights}
    \item Introduces FP-IRL: Fokker--Planck inverse reinforcement learning
    \item Infers reward and transition functions without needing known transition dynamics
    \item Establishes and exploits a phenomenological connection between the gradient structure of a class of continuous-space Markov Decision Processes and the associated Fokker--Planck equation
    \item Achieves accurate recovery of agent behavior with physical interpretability
    \item Demonstrates convergence and efficiency on synthetic and benchmark problems
\end{highlights}

\end{frontmatter}

\section{Introduction}

Many complex dynamical systems, ranging from cancer cell migration and human decision-making to crowd behavior, are composed of autonomous agents interacting with uncertain environments. These agents often make decisions in response to latent, unobserved incentives and operate under significant heterogeneity and stochasticity. 
Understanding such systems is challenging: traditional mechanistic models based on ordinary or partial differential equations (ODEs, PDEs) typically capture population-level dynamics, but struggle to account for goal-directed, agent-level decision behavior, especially when the governing principles are unknown or unobservable.

In such settings, \textit{Markov decision processes} (MDPs)~\cite{Puterman1994,Sigaud2010} provide a powerful modeling framework that explicitly represents individual decision-making under uncertainty.
When the reward structure driving agent behavior is unknown, \textit{inverse reinforcement learning} (IRL)~\cite{russell1998learning,ng2000algorithms,ratliff2006maximum,ramachandran2007bayesian,ziebart2008maximum,fu2018learning} offers a principled, data-driven approach to recover it from observed behavior.
The central idea of IRL is to infer a reward function such that an optimal policy under this reward would explain the observed agent's trajectories. This paradigm has been successfully applied in domains such as robotics~\cite{levine2012continuous,finn2016guided}, human behavior modeling~\cite{ratliff2006maximum,ziebart2008maximum,Hossain2022Bayesian}, and biology~\cite{Kalantari2020Unreasonable}, and has inspired a wide variety of algorithmic developments, including
maximum margin methods~\cite{ng2000algorithms,ratliff2006maximum}, %
feature matching~\cite{abbeel2004apprenticeship},
entropy-regularized IRL 
\cite{ziebart2008maximum,ziebart2010modeling,ziebart2010thesis},
adversarial IRL ~\cite{fu2018learning,yu2019multi,Henderson2018OptionGAN}, 
Bayesian IRL~\cite{ramachandran2007bayesian,Kalantari2020Unreasonable},
and offline IRL~\cite{zeng2023understanding,yue2023clare}.
See~\citet{Arora2021IRLSurvey,Adams2022IRLSurvey} for comprehensive surveys.

Despite this progress, two major challenges persist in IRL, especially in scientific applications.
First, most IRL algorithms assume access to or require empirical estimation of the environment's transition dynamics, which may not be feasible in systems where transitions are unknown or unobservable. For instance, in cancer biology, the rules governing how cells respond to local cues are poorly understood and inaccessible to direct sampling. 
Second, IRL methods relying on deep neural networks~\cite{herman2016inverse,yue2023clare} often lack interpretability, limiting their ability to generate meaningful scientific hypotheses or insights into the system's underlying mechanisms.

At the same time, many natural and engineered systems
are known to follow mechanistic laws, such as those described by 
stochastic differential equations (SDEs) and their continuum limits: the \textit{Fokker--Planck} (FP) PDEs~\cite{risken1996fokker}.
These laws capture important physical structure, including conservation, drift, and diffusion. Importantly, they describe how population-level densities evolve, not how individual agents make decisions. 
We seek to bridge this gap between physics-based population-level modeling and decision-centric, agent-based modeling.

In this work, we propose a novel framework: \textit{Fokker--Planck inverse reinforcement learning} (FP-IRL). 
Our key insight is a \textbf{correspondence between FP dynamics and a class of physics-based MDPs}, which allows us to recast IRL as a regression problem constrained by FP physics. This formulation enables us to:
\begin{itemize}
\item infer both the transition and reward functions without sampling the environment; 
\item preserve interpretability through physically meaningful quantities (e.g., drift and diffusion); and 
\item avoid nested policy optimization by leveraging variational system identification (VSI)~\cite{Wang2019Variational,Wang2021Variational} to infer governing PDEs.
\end{itemize}
We develop the FP-IRL algorithm based on this insight and validate it on both synthetic examples and a modified version of the classic Mountain Car benchmark, redesigned to follow FP dynamics. Our results demonstrate accurate recovery of reward, transition, and policy functions, along with empirical convergence under mesh refinement.

The paper is organized as follows.
\Cref{sec:preliminaries} introduces relevant IRL background and problem formulation. 
\Cref{sec:fpirl} presents our FP-IRL framework and the connection between FP dynamics and MDPs. 
\Cref{sec:vsi} details the VSI method. 
\Cref{sec:experiments} demonstrates results on numerical examples. 
\Cref{sec:discussion} discusses the broader significance and limitations of our approach.
\Cref{sec:conclusion} concludes the paper with a summary of key findings.

\section{Problem Formulation}
\label{sec:preliminaries}

\subsection{Preliminaries}

\begin{figure}[htbp]
    \centering
    \includegraphics[width=0.7\textwidth]{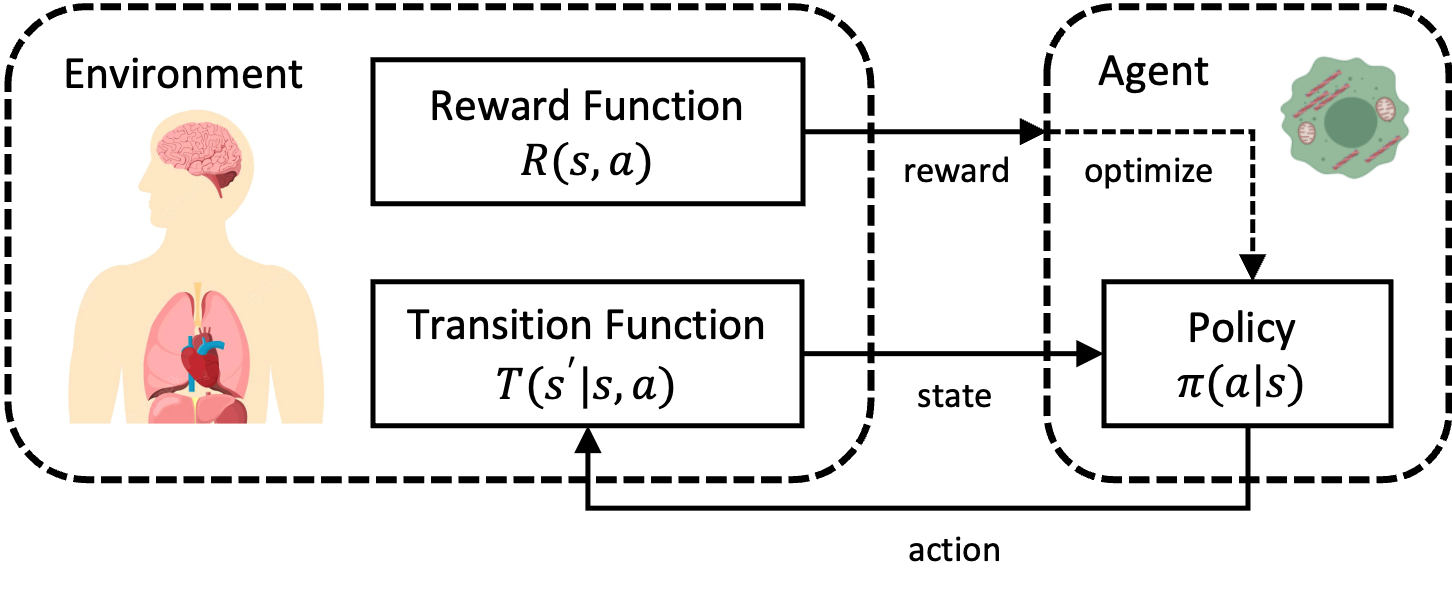}
    \caption{Schematic illustration of an agent's iterative interaction with the environment, modeled as an MDP.}
    \label{fig:mdp}
\end{figure}

We consider an MDP defined by a tuple
$ \CalM \triangleq \{ \CalS, \CalA, \rho_0(\cdot), \allowbreak R(\cdot), T(\cdot) \}$, where 
\begin{itemize}
\item $\CalS \subseteq \RR^{d_s}$ is the state space
with states $\bss \in \CalS$, 
\item $\CalA \subseteq \RR^{d_a}$ is the action space
with actions $\bsa \in \CalA$, 
\item $\rho_0(\bss): \CalS \rightarrow \RR^+$ is the initial state distribution,
\item $R(\bss, \bsa): \CalS \times \CalA \rightarrow \RR$ is the reward function, and
\item $T(\bss' | \bss, \bsa): \CalS \times \CalS \times \CalA \rightarrow \RR^+$ is the state transition probability function, which gives the probability of transitioning to state $\bss'$ when taking action $\bsa$ in state $\bss$. 
\end{itemize}
An agent interacts with the environment by following a stochastic policy $\pi(\bsa | \bss): \CalS \times \CalA \rightarrow \RR^+$,  
which specifies the probability of taking action $\bsa$ in state $\bss$. At each discrete time step, the agent samples an action from $\pi$, receives a reward, and transitions to a new state according to $T$ (see \cref{fig:mdp}).
While this formulation adopts a discrete-time perspective, we later consider its continuous-time analogue, where state transitions are governed by stochastic diffusion dynamics.

A central object of interest in an MDP is the \textit{state-action value function} (or Q-function), $Q^{\pi}(\bss, \bsa): \CalS \times \CalA \rightarrow \RR$, which 
evaluates the expected cumulative reward obtained by starting from state $\bss$, taking action $\bsa$, and subsequently following policy $\pi$. In the infinite-horizon discounted setting, it is defined as:
\begin{align}
        \label{eq:culmulative_reward}
        Q^{\pi}(\bss, \bsa) &=  \EE_{\subalign{\bss_t &\sim T(\cdot | \bss_{t-1}, \bsa_{t-1}) \\ \bsa_t &\sim \pi(\cdot | \bss_t)}} \[\sum^{\infty}_{t=k} \gamma^{t-k} R(\bss_t, \bsa_t) \bigg| \bss_k = \bss, \bsa_k =\bsa \],
\end{align}
where $\gamma\in [0, 1)$ is the discount factor, used to down-weight future rewards.
The Q-function satisfies the \textit{Bellman expectation equations}~\cite{bellman1952theory}:
\begin{align}
    \label{eq:bellman_expectation_q}
    Q^{\pi}(\bss, \bsa) &= R(\bss, \bsa) + \gamma \EE_{\bss' \sim T(\cdot| \bss, \bsa)} \big[ V^{\pi}(\bss') \big], \\
    \label{eq:bellman_expectation_v}
    V^{\pi}(\bss) &= \EE_{\bsa \sim \pi(\cdot | \bss)} \big[Q^{\pi}(\bss, \bsa)  \big],
\end{align}
where $V^{\pi}(\bss)$ is the \textit{state value function} (or V-function), which represents the expected cumulative reward when starting at state $\bss$ and following policy $\pi$ thereafter. 

\textit{Reinforcement learning} (RL), as illustrated in \cref{fig:rl}, aims to find an optimal policy $\pi^{\ast}$
that
maximizes the expected cumulative discounted reward (also known as the expected return):
\begin{align}
    \label{eq:optimal_policy_objective_reward}
    \pi^{\ast} &= \argmax_{\pi \in \Pi} \quad \EE_{\subalign{\bss_{0} &\sim \rho_{0}(\cdot) \\ \bss_{t} &\sim T(\cdot | \bss_{t-1}, \bsa_{t-1}) \\ \bsa_t &\sim \pi(\cdot | \bss_t) }}\[ \sum_{t=0}^{\infty} \gamma^{t} R(\bss_t, \bsa_t)\]\\
    \label{eq:optimal_policy_objective_value}
    &= \argmax_{\pi \in \Pi} \quad \EE_{\subalign{\bss_0 &\sim \rho_0(\cdot) \\ \bsa_0 &\sim \pi(\cdot | \bss_0)}} \big[ Q^{\pi}(\bss_0, \bsa_0) \big],
\end{align}
where $\Pi$ denotes the space of admissible policies, assumed here to be time-invariant and memoryless.

\textit{Inverse reinforcement learning}, shown in \cref{fig:irl}, addresses the inverse problem: given the observed behavior of an expert agent, the goal is to recover the underlying reward function $R$ that explains the observed behavior.
In many settings, such as modeling biological agents or human decision-making, explicitly specifying a reward function is challenging. IRL offers a data-driven approach to infer the agent's implicit objectives directly from observed trajectories.
The input to IRL consists of expert trajectories 
$ \CalD = \allowbreak \big \{ \big(\bss_0^{(i)}, \bsa_0^{(i)}, \allowbreak  \cdots, \bss_{\tau_i}^{(i)}, \bsa_{\tau_i}^{(i)} \big) \big \}_{i=1}^{m}$, where $m$ is the number of trajectories and $\tau_i$ the length of the $i$-{th} trajectory. These trajectories are assumed to be generated by an expert following a (near-) optimal policy with respect to some unknown reward function $R$. 

In classical IRL, only the reward function $R$ is unknown; all other components of the MDP---particularly the transition function $T$---are assumed to be known \textit{a priori} or empirically estimated from data. 
This knowledge of the transition dynamics is essential, as it enables the computation of optimal policies for candidate rewards and the simulation of new trajectories.
This allows IRL algorithms to iteratively adjust $R$ so as to reduce the discrepancy between simulated and observed behaviors.

\subsection{Problem statement: IRL with physics-constrained transition inference}
\label{sec:problem_statement}
In many real-world scenarios (e.g., biological or human systems), not only is the reward function $R$ unknown, but the transition function $T$ is also unobserved.
In such cases, we do not have access to an environment or simulator for sampling from $T$.
The absence of $T$ introduces a fundamental indeterminacy: many distinct reward-transition pairs may be equally consistent with the observed behavior, exacerbating the ill-posed nature of the IRL problem. 

To address this, 
offline IRL approaches~\cite{zeng2023understanding,yue2023clare} 
typically estimate the transition function empirically from data before inferring the reward.
\citet{herman2016inverse} proposes a purely data-driven approach to jointly infer both reward and transition using neural networks. However, these methods do not incorporate any known physical structure into the transition dynamics, 
making them more susceptible to overfitting and less amenable to scientific interpretation.

We consider a particular class of MDPs whose policy-induced Markov process over the state-action pairs can be represented by a FP PDE.
In this class of MDPs, we further wish to treat cases in which the Q-function is equivalent to the negative of the potential function that governs the drift of the FP dynamics. In this, we are motivated by the maximization of the Q-function and the drift down the gradient of the potential function, corresponding to its minimization. For this reason, we assume the equivalence of the Q-function to the negative potential. Entropy regularization plays a further role, which we discuss in \cref{sec:agent_policy}.

\begin{figure}[htbp]
    \centering
    \begin{subfigure}[c]{0.32\textwidth}
        \centering
        \includegraphics{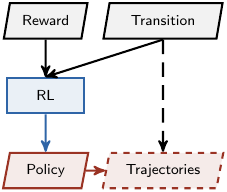}
        \caption{RL}
        \label{fig:rl}
    \end{subfigure}
    \hfill
    \begin{subfigure}[c]{0.32\textwidth}
        \centering
        \includegraphics{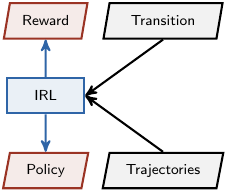}
        \caption{IRL}
        \label{fig:irl}
    \end{subfigure}
    \hfill
    \begin{subfigure}[c]{0.32\textwidth}
        \centering
        \includegraphics{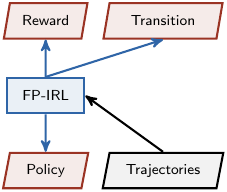}
        \caption{FP-IRL}
        \label{fig:fpirl}
    \end{subfigure}
    \caption{Comparison of the objectives of RL, IRL, and FP-IRL. 
    (a) RL learns an optimal policy given {known reward and transition functions in} an MDP. Using the learned policy, one can generate trajectories by interacting with the environment.
    {The dashed arrow represents the indirect output (trajectories) of the algorithm.}
    (b) IRL infers the reward function {and corresponding policy} from observed expert trajectories, assuming access to known transition dynamics.
    (c) FP-IRL extends IRL by simultaneously inferring both the reward and transition functions, with the latter constrained by physical principles.
    In all subfigures, black and red parallelograms denote inputs and outputs, respectively, while blue 
    rectangles represent algorithmic component. }
    \label{fig:comparison_rl_irl_fpirl}
\end{figure}

We propose \textbf{FP-IRL}, a novel physics-constrained framework for IRL. FP-IRL leverages the FP PDE to model the evolution of continuous state-action distributions, enabling the simultaneous inference of both reward and transition functions in a manner consistent with underlying physical laws  (see \cref{fig:fpirl} for comparison with RL and classical IRL). 
This is particularly important in systems with continuous, stochastic dynamics, where transitions follow diffusive behavior governed by physical constraints.
By embedding this structure into the learning process, FP-IRL regularizes the ill-posed IRL problem and improves interpretability.
Additional benefits of incorporating physics-based constraints are discussed in \cref{sec:discussion_significance}.

\section{Fokker--Planck Inverse Reinforcement Learning}
\label{sec:fpirl}

\begin{figure}[htbp]
    \centering
    \includegraphics[width=\textwidth]{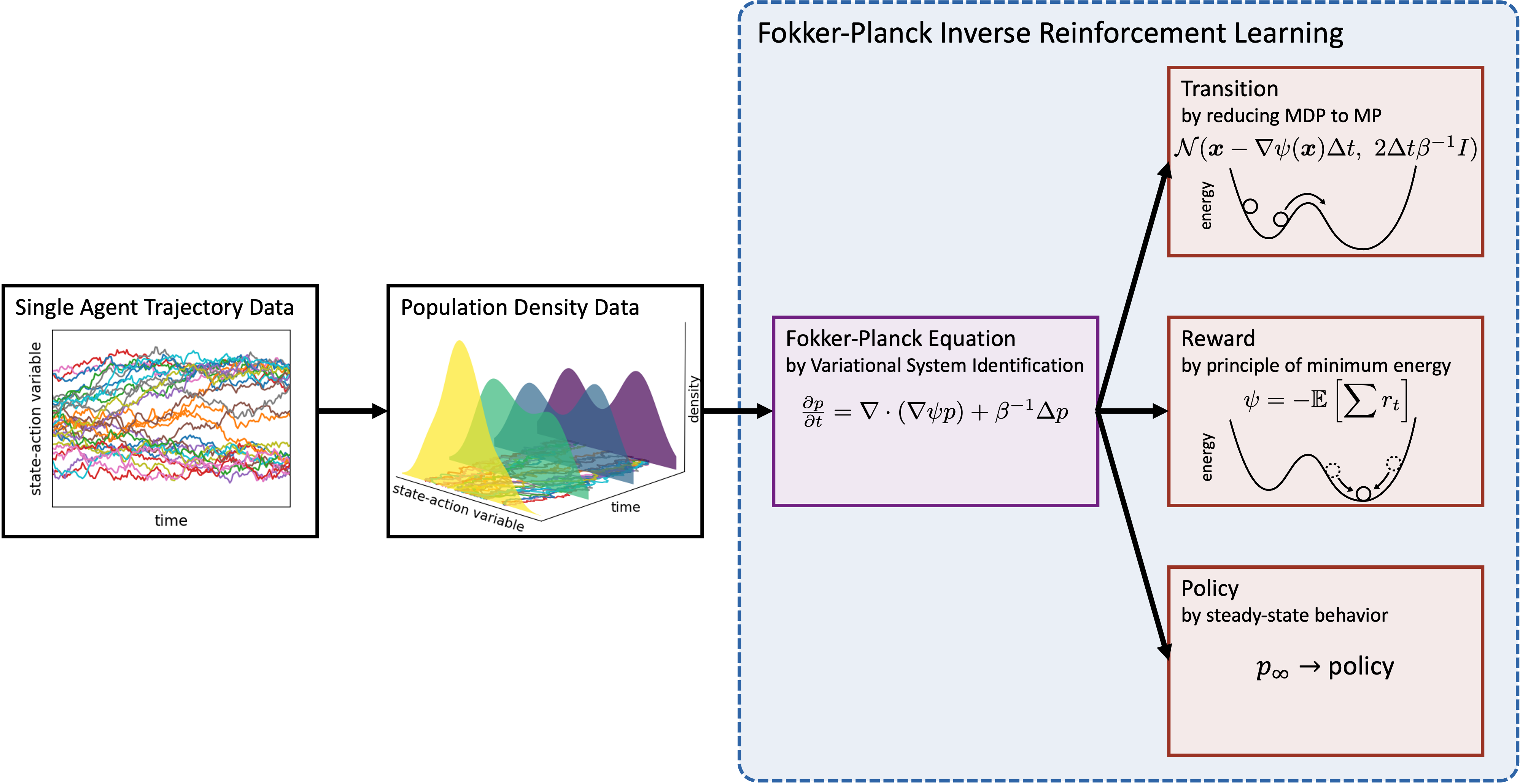}
    \caption{Schematic overview of the FP-IRL framework, which infers both reward and transition functions by leveraging the evolution of state-action densities under FP dynamics.}
    \label{fig:fpirl_summary}
\end{figure}

In this section, we present FP-IRL, a physics-constrained framework for IRL (see \cref{fig:fpirl_summary}).
We begin by formulating the transition dynamics of the MDP using the FP PDE, which describes the time evolution of state-action distributions under stochastic diffusion.
Building on this formulation, we make a conjecture on an equivalence
between the FP PDE and the MDP, grounded in a minimum energy principle. This connection enables the joint estimation of the transition function, reward function, and policy from observed data. 

\subsection{Fokker--Planck physics for learning the transition function}
\label{sec:physics_modeling}
The FP PDE arises in a wide range of physical systems where the time evolution of a probability density function is governed by a transport process. This equation provides a natural framework for modeling the dynamics of physical and biological systems that exhibit continuous, stochastic behavior~\cite{risken1996fokker}. Motivated by this, we incorporate physics into IRL by learning the transition dynamics of an MDP through the FP evolution of probability density functions.

\begin{figure}[htbp]
    \begin{subfigure}{\textwidth}
        \centering
        \includegraphics{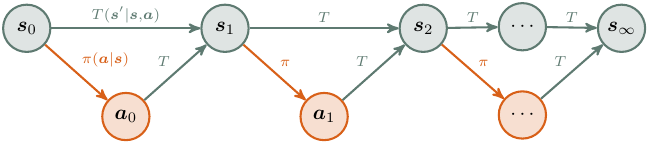}
        \caption{State-action evolution under a fixed policy in an MDP.}
    \end{subfigure}
    \begin{subfigure}{\textwidth}
        \centering
        \includegraphics{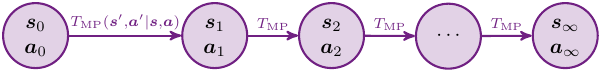}
        \caption{State evolution in the corresponding MP induced by the MDP and policy.}
    \end{subfigure}
    \caption{Illustration of how an MDP under a fixed policy induces a MP over the lumped state-action variable $\bsx=(\bss,\bsa)$, with transitions governed by the joint dynamics.}
    \label{fig:mdp2mp}
\end{figure}

We begin by noting that an 
MDP under a 
fixed 
(time-invariant)
policy $\pi$ induces a \textit{Markov process} (MP) over the lumped state variable 
$\bsx = (\bss, \bsa) \in \Omega$, where $\Omega = \CalS \times \CalA \subseteq \RR^{d}$ (see \cref{fig:mdp2mp}).
The corresponding MP transition function is given by:
\begin{align}
    \label{eq:joint_transition}
    T_{\text{MP}} \big(\bsx' | \bsx \big)= T_{\text{MP}} \big( \bss', \bsa' | \bss, \bsa \big) = \pi \big(\bsa'|\bss' \big) T \big(\bss' | \bss, \bsa \big),
\end{align}
where the Markov property (i.e., memoryless) implies that $\pi(\bsa'|\bss')$ is independent of the previous state-action pair $(\bss,\bsa)$.
Given $T_{\text{MP}}$, the original MDP transition can be recovered via marginalization:
\begin{align}
    \label{eq:mdp_transition}
    T \big(\bss' | \bss, \bsa \big) = \int_{\CalA} T_{\text{MP}} \big(\bss', \bsa' | \bss, \bsa \big) \, \mathrm{d} \bsa'.
\end{align}

We frame the problem of learning the MDP transition function as one of inferring the corresponding MP transition from 
observed data.
To do so, we leverage the connection between MPs and \textit{stochastic differential equations} (SDEs). 
In particular, we assume the dynamics of the lumped state $\bsx(t)$ are governed by an It\^{o} SDE. This class of
equations is broadly applicable in settings where agents are influenced by both directed forces (e.g., goal-seeking behavior) and random perturbations (e.g., environmental noise), such as in chemotaxis of cells, swarm behavior, or social navigation (see \cref{sec:discussion_significance}). The governing It\^{o} SDE is given by:
\begin{align}
    \label{eqn:SDE}
    \mathrm{d} \bsx(t) = -\nabla \psi \big( \bsx(t) \big) \, \mathrm{d} t + \sqrt{2 \beta^{-1}} \, \mathrm{d} \bsw(t),
\end{align}
where 
$\psi(\bsx): \Omega \rightarrow \RR$ is a potential function, $\beta$ is an inverse temperature parameter from statistical physics, and $\bsw(t)$ is a standard $d$-dimensional Wiener process. This SDE captures two competing effects: deterministic drift down the potential gradient $-\nabla \psi$, and stochastic diffusion via Brownian motion.

For an infinitesimal time step $\Delta t$, the resulting transition distribution for this process is Gaussian up to first-order approximation of $\psi$~\cite{risken1996fokker,bressloff2014stochastic}:
\begin{align}
    \label{eq:gaussian_transition}
    T_{\text{MP}}(\bsx' | \bsx) = \left(\frac{\beta}{4 \pi \Delta t}\right)^{d/2} \exp{\(\frac{-\beta||\bsx' - \bsx + \nabla \psi(\bsx) \Delta t ||^2}{4  \Delta t}   \)}.
\end{align}
Thus, characterizing the MP transition amounts to estimating the potential function $\psi$ and the inverse temperature $\beta$.

Although it is theoretically possible to infer the parameters $\psi$ and $\beta$ directly from the SDE, doing so is often computationally intensive and highly sensitive to trajectory-level noise. 
The SDE describes the stochastic evolution of single-agent sample paths, which can fluctuate significantly across realizations. In contrast, the corresponding FP PDE governs the time evolution of the probability density $p(\bsx, t)$,
offering a macroscopic perspective that captures population-level dynamics. 
This perspective smooths over individual randomness, improves robustness to noise, and enables parameter inference directly at the level of distributions without the need to simulate or regress over individual trajectories.
We therefore adopt the FP formulation:
\begin{align}
    \label{eq:fokker_planck_eq}
    \pp{p(\bsx, t)}{t} = \nabla \cdot \big( p(\bsx, t) \nabla \psi(\bsx) \big) + \beta^{-1} \nabla^2 p(\bsx, t).
\end{align}
This PDE form allows us to leverage established tools from the inverse problem literature, particularly \textit{variational system identification} (VSI), which we describe in \cref{sec:vsi}, to infer $\psi$ and $\beta$ 
from the probability densities.

\subsection{Free energy and its connection to the Q-function in physics-based MDPs}
\label{sec:free_energy}
Having established the transition dynamics of the MDP through \cref{eq:mdp_transition}, 
the remaining challenge in IRL is to estimate the reward function and the corresponding optimal policy. 
This sets the stage for a modeling assumption used by FP-IRL: for the class of physics-based MDP whose induced state-action dynamics admit an FP description, the Q-function is structurally equivalent to the negative potential function in the FP PDE of the MDP-induced MP. This equivalence naturally leads to the introduction of a free energy functional that governs the evolution of the system.

\subsubsection{Free energy in statistical mechanics}
In statistical mechanics, the free energy functional is fundamental for characterizing equilibrium behavior.
It reflects a balance between internal energy, represented by a potential energy function $\psi$, and system disorder, measured by the differential entropy
\begin{align}
    H_{\bsx}(p) = - \int_{\Omega} p(\bsx) \log p(\bsx) \, \mathrm{d} \bsx.
\end{align}
For a probability density function $p$ and a potential $\psi$, the \textit{free energy} is defined as: 
\begin{align}
    \label{eq:free_energy}
    F(p, \psi) =& \int_{\Omega} \psi(\bsx) p(\bsx) \, \mathrm{d} \bsx - \beta^{-1} H_{\bsx}(p).
\end{align}
For notational simplicity, we omit the explicit time dependence unless otherwise noted (writing $p_t(\bsx)$ when needed).
According to the \textit{principle of minimum free energy}, a stochastic system governed by FP dynamics 
evolves toward an equilibrium distribution $p_{\infty}(\bsx)$ that minimizes $F(p,\psi)$, with the unique minimizer given by the Gibbs--Boltzmann distribution~\cite{risken1996fokker, Jordan1998Variational}.

\citet{Jordan1996Free,Jordan1998Variational} further formalized FP dynamics as a {Wasserstein} gradient flow, showing that the discrete-time update
\begin{align}
    \label{eq:free_energy_fokker_planck}
    p_{t_{k+1}} = \argmin_{p} W_2(p_{t_k}, p)^2 + \Delta t \, F(p, \psi)
\end{align}
converges to the solution of the FP PDE as $\Delta t \to 0$,
where $W_2(\cdot)$ denotes the Wasserstein-2 distance.
Here, the time evolution of $p$ is described as a sequence of minimization problems. 
At each step, minimizing the free energy functional is regularized by a transport cost, measured by the Wasserstein distance from the previous state distribution. 
This Wasserstein distance acts as a movement limiter: it penalizes large, non-physical shifts in the distribution and thus enforces smooth and continuous evolution over time. 
Without this regularization, this optimization problem would lose its dependence on the time step, and dictate the minimizer of the free energy as the single attainable solution at all times, resulting in an instantaneous transition to the free energy minimizer.

\subsubsection{Free energy in physics-based MDPs}
This variational framework for FP dynamics has a compelling analogue in MDPs. 
In MDPs, an agent's optimal policy seeks to maximize the expected cumulative reward (i.e., value), subject to the stochastic transition dynamics of the environment (see \cref{eq:optimal_policy_objective_reward}). 
The optimal policy thus leads the agent toward regions of high value, which is analogous (inversely) to the role of low potential energy in an FP system, balanced by system entropy.
Furthermore, in physics-based MDPs, the environmental transitions are typically continuous and smooth, reflecting physical constraints that prevent abrupt changes in state.
This smoothness requirement parallels the effect of Wasserstein regularization in \cref{eq:free_energy_fokker_planck}, which forces the evolution of the probability density to remain bounded over time.

This observation raises a natural question: can the optimization behavior in an physics-based MDP---typically framed as value function maximization 
---be reinterpreted through the lens of free energy minimization?
If so, this connection would offer both a theoretical foundation for physics-constrained IRL and a practical regularizer for addressing the ill-posed nature of inverse problems. This motivates the following modeling assumption.
\begin{assumption}[Value-Potential Equivalence]
    \label{thm:equivalence}
    The Q-function in a physics-based MDP is equivalent to the negative potential function in the corresponding FP system:
    \begin{align}
        \label{eq:conjecture_equivalence}
        Q^{\pi}(\bss, \bsa) = -\psi(\bsx), \quad \text{where } \bsx = (\bss, \bsa).
    \end{align}
\end{assumption}
This assumption implies a structural similarity between the FP evolution of distributions and the dynamics of decision-making in MDPs. Under this equivalence, the free energy functional becomes a bridge between the probabilistic evolution of physical systems and the value-driven optimization in RL. 

To further elucidate this equivalence and facilitate analysis, we derive the explicit form of the free energy functional in the MDP setting under \cref{thm:equivalence}.
The joint state-action distribution can be written as $p(\bss, \bsa) = \rho(\bss) \pi(\bsa | \bss)$.
By the \textit{chain rule for differential entropy}~\cite[Theorem~2.2.1]{cover2012elements}, the joint entropy decomposes as $H_{\bss, \bsa}(p) = H_{\bss}(\rho) + H_{\bsa | \bss} (\pi)$, where $H_{\bss}(\rho)$ is the entropy of the marginal state distribution $\rho$, and $H_{\bsa | \bss}(\pi) = \EE_{\bss \sim \rho} \big[ H_{\bsa} \big(\pi(\cdot | \bss) \big) \big]$ is the conditional entropy, representing the expected entropy of the policy over the states.
Substituting these expressions into the free energy functional in \cref{eq:free_energy}, and using the value-potential equivalence from \cref{thm:equivalence}, we obtain the free energy of the physics-based MDP:
\begin{align}
    \label{eq:free_energy_mdp}
    F(\rho, \pi) 
    =& - \int_{\CalS} \rho(\bss) \int_{\CalA} \pi(\bsa | \bss) Q^{\pi}(\bss, \bsa) \, \mathrm{d} \bsa \, \mathrm{d} \bss - \beta^{-1} \( H_{\bss}(\rho) + H_{\bsa | \bss} (\pi) \).
\end{align}
Here, the free energy in the MDP setting is fully characterized by the pair $(\rho, \pi)$, without requiring explicit dependence on $Q^{\pi}$, 
as $Q^{\pi}$ is uniquely determined by $\pi$.

The minimization of the free energy in MDPs (\cref{eq:free_energy_mdp}) can be understood as a two-step process. 
\begin{itemize}
\item \textit{Policy optimization}: for any $\rho$, minimizing $F(\rho, \pi)$ with respect to $\pi$ yields the optimal policy $\pi^{\ast}$, whose value function $Q^{\pi^{\ast}}$ defines the lowest possible potential energy landscape, $\psi = -Q^\pi$.
\item \textit{Distributional evolution}:
by applying $\pi^{\ast}$ over time, the system evolves toward its equilibrium where free energy is further minimized with respect to $\rho_t$ over time, in alignment with the FP evolution toward equilibrium.
\end{itemize}
In IRL, where $\pi$ is time-invariant and assumed optimal, the agent repeatedly applies $\pi^{\ast}$, and the IRL problem becomes inferring the underlying FP potential function from observed behavior. 

Through this assumption, we establish a novel connection between the FP PDE---a foundational model in statistical physics---and the MDP formalism underlying sequential decision-making. This connection offers both a conceptual bridge and a practical regularization strategy in IRL, helping to mitigate the ill-posedness inherent in simultaneously recovering transitions and rewards from observed behavior.
A more detailed discussion of the implications of this add{assumption} is provided in \cref{sec:discussion_significance}.

In the following sections, we build on this add{assumption} to infer the governing FP dynamics and to recover the underlying policy and reward functions from observed data.

\subsubsection{Empirical demonstration of free energy minimization in an MDP}
\begin{figure}[htbp]
    \centering
    \includegraphics[width=0.5\textwidth]{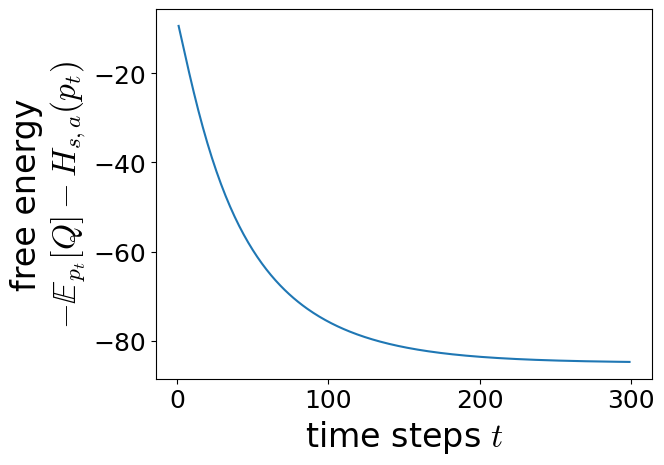}
    \caption{Empirical validation of the free energy principle in an MDP setting. In the grid world environment, the agent's state-action distribution evolves toward a equilibrium that minimizes the free energy, consistent with \cref{thm:equivalence}. 
    }
    \label{fig:free_energy_illustration}
\end{figure}

To further illustrate that the principle of minimum free energy holds in MDPs, we examine a synthetic Grid World environment (see \cref{sec:synthetic_example} for details).
The agent navigates a bounded two-dimensional state space, with state $\bss = (x, y) \in [-1,1]^2$, by selecting continuous velocity actions $\bsa = (v_x, v_y)$ to reach a designated goal tile.
The reward function $R(\bss, \bsa) = |x| - y - \sqrt{v_x^2 + v_y^2}$ encourages movement toward lower corners of the grid while penalizing high velocities.
State transitions follow a Gaussian distribution: $\[x' \ y'\]^{\top} \sim \CalN \big(\[x \ y\]^{\top} + \[v_x \ v_y\]^{\top} \Delta t, \sigma^2 I \big)$.
Using this setup, we simulate the evolution of the agent's state-action distribution $p(\bss, \bsa)$ under the optimal policy $\pi^{\ast}$ and track its free energy (in \cref{eq:free_energy_mdp}) over time.
As shown in \cref{fig:free_energy_illustration}, the free energy decreases monotonically over time and converges toward a minimum at equilibrium. This empirical result supports the hypothesis that optimal decision-making in an MDP can be interpreted as a process of free energy minimization, thus reinforcing the validity of \cref{thm:equivalence}.

\subsection{Optimal policy constrained by FP dynamics}
\label{sec:agent_policy}
In this section, we derive the agent's optimal policy under FP-constrained MDP dynamics. 
At equilibrium, minimizing the free energy yields the Boltzmann policy; during transient evolution, the agent's policy reflects a balance of smoothness and optimality.

\subsubsection{Equilibrium case: free energy minimization and the Boltzmann policy} 
The principle of minimum free energy establishes a foundational conceptual connection between statistical mechanics and RL.
In particular, the equilibrium distribution $p_\infty$ of a stochastic system minimizes the free energy functional, linking physical equilibria to optimal policies in MDPs.

As discussed in \cref{sec:free_energy}, the equilibrium distribution $p_{\infty}$ minimizes the free energy functional $F(p,\psi)$ (in \cref{eq:free_energy}), and this minimizer takes the form of a Gibbs--Boltzmann distribution~\cite{risken1996fokker, Jordan1998Variational}:
\begin{align}
    p_{\infty}(\bss, \bsa) =
    Z^{-1} \exp \big(- \beta \psi(\bss, \bsa) \big),
\end{align} 
where the normalization constant is given by $Z = \int_{\CalS} \int_{\CalA} \exp \big(- \beta \psi(\bss, \bsa) \big) \, \mathrm{d} \bsa \, \mathrm{d} \bss$. The corresponding marginal state distribution is given by:
\begin{align}
    \rho_{\infty}(\bss) = \int_{\CalA} p_{\infty}(\bss, \bsa) \, \mathrm{d} \bsa 
    = Z^{-1} \int_{\CalA} \exp \big(-\beta \psi(\bss, \bsa) \big) \, \mathrm{d}\bsa,
\end{align}
which induces the conditional distribution over actions:
\begin{align}
    \label{eq:conditional_steadystate}
    p_{\infty}(\bsa | \bss) = \frac{p_{\infty}(\bss, \bsa)}{\rho_{\infty}(\bss)}
    = \frac{\exp \big( - \beta \psi(\bss, \bsa) \big)}{\int_{\CalA} \exp \big( - \beta \psi(\bss, \bsa') \big) \, \mathrm{d}\bsa'}.    
\end{align}

For the physics-based MDP, recalling the decomposition of the joint distribution $p(\bss, \bsa) = \rho(\bss) \pi(\bsa | \bss)$, 
an equivalent result can be derived by directly minimizing the free energy functional (\cref{eq:free_energy_mdp}) with respect to the policy $\pi$, for any given state distribution $\rho$. Treating $H_{\bss}$ as a constant with respect to $\pi$, this objective simplifies to:
\begin{align}
    \label{eq:free_energy_rl_objective_expand}
    \argmin_{\pi \in \Pi}  \int_{\CalS} \rho(\bss) \int_{\CalA} \pi(\bsa | \bss) \[ -Q^{\pi}(\bss, \bsa) + \beta^{-1} \log \pi(\bsa | \bss) \] \mathrm{d} \bsa \, \mathrm{d} \bss,
\end{align}
with the optimal solution $\pi^{\ast}(\bsa | \bss)$ given by:
\begin{align}
    \label{eq:boltzmann_policy}
    \pi^{\ast}(\bsa|\bss) 
    = \frac{\exp \big( \beta Q^{\pi}(\bss, \bsa) \big)}{\int_{\CalA} \exp \big( \beta Q^{\pi}(\bss, \bsa') \big) \, \mathrm{d}\bsa'}.
\end{align}

The expression in \cref{eq:conditional_steadystate,eq:boltzmann_policy} coincides with the Boltzmann policy widely used in entropy-regularized RL and IRL~\cite{sallans2004reinforcement,ziebart2010modeling,haarnoja2018soft,skalse2023misspecification}. 
This match provides further support for \cref{thm:equivalence} and its implications for policy recovery in FP-IRL, particularly in the equilibrium regime.

\subsubsection{Transient case: variational policy optimization and movement limitation via Wasserstein regularization}
\label{sec:transientWassreg}
In \cref{sec:physics_modeling}, we showed that an MDP governed by a time-invariant policy induces an MP over the joint state-action space $\Omega$.  
We now ``lift'' the MP back to an MDP by interpreting the policy $\pi$ during transient dynamics through the discrete-time optimization formulation given in \cref{eq:free_energy_fokker_planck}.
This clarifies how distributional evolution can be interpreted as balancing (i) a transport-based movement limitation, encoded by the Wasserstein term, and (ii) a free-energy objective that reflects value-seeking behavior tempered by entropy.
Our goal is not to treat \cref{eq:free_energy_fokker_planck} as a computational primitive for FP-IRL, but rather as an interpretive lens for the structure of physics-constrained decision processes.

Analyzing the upper bound of the minimization problem in \cref{eq:free_energy_fokker_planck}, we see that $p_{t_{k+1}}$ is the minimizer of an energy that is itself bounded from above according to a ``triangle inequality-like'' property of Wasserstein-2 distance (see Appendix \ref{appsec:wasserstein_chain_rule} or~\citet{Chemseddine2025Conditional} for details):
\begin{align}
\begin{split}
    & \min_{p} W_2^2\(p_{t_k}, p \) + \Delta t \, F(p) \\
    \le & \min_{\rho, \pi} W_2^2\(\rho_{t_k}, \rho \) 
    + \EE_{(\bss_{t_k}, \bss) \sim \gamma_s^{\ast}} \[ W_2^2 \big(\pi(\cdot|\bss_{t_k}), \pi(\cdot|\bss) \big) \] + \Delta t \, F(\rho, \pi),
\end{split}
\label{eq:comp-obj}
\end{align}
for $p(\bss, \bsa) = \rho(\bss) \pi(\bsa |\bss)$. 
Similar to the regularizing effect of $W_2(p_{t_k}, p)$ in \cref{eq:free_energy_fokker_planck}, 
the first term $W_2^2 (\rho_{t_k}, \rho )$ penalizes large deviations in the state distribution,
while the second term $\EE_{(\bss_{t_k}, \bss) \sim \gamma_s^{\ast}} \big[ W_2^2 \big( \pi(\cdot|\bss_{t_k}), \pi(\cdot|\bss) \big) \big]$ encourages smooth changes in the policy across states that are likely to be reached in subsequent steps.
The free energy term serves as an objective for policy optimization, balancing expected return and policy entropy, as in \cref{eq:free_energy_rl_objective_expand}. 
Since the dynamics are subject to a Wasserstein flow by the optimality condition in \cref{eq:free_energy_fokker_planck}, the ``triangle inequality'' of  Wasserstein-2 distance guarantees that the ``energy'' attained by the minimizer of the joint distribution $p(\bss_t,\bsa_t)$ bounds from below the composite objective  \cref{eq:comp-obj} that itself controls the temporal variations of the state distribution and policy.
Altogether, this framework unifies control over movement of density and policy between discrete time steps and optimality in physics-constrained MDP policies.

\subsection{Inverse Bellman equation}
\label{sec:inverse_bellman}
Given the transition function $T$ defined in \cref{eq:mdp_transition}, the Q-function $Q^{\pi}$ in \cref{thm:equivalence}, and the policy $\pi$ in \cref{eq:free_energy_rl_objective_expand}, 
all obtained through the FP PDE as discussed in \cref{sec:physics_modeling,sec:free_energy,sec:agent_policy}, the reward function $R$ can be recovered via the inverse Bellman equation:
\begin{align}
    \label{eq:inverse_bellman}
    R(\bss, \bsa) = Q^{\pi}(\bss, \bsa) - \gamma \EE_{\subalign{\bss' &\sim T(\cdot | \bss, \bsa) \\ \bsa' &\sim \pi(\cdot | \bss')}} \big[ Q^{\pi}(\bss', \bsa') \big].
\end{align}
This suggests that, for a given transition kernel and value function, there exists a unique reward function, as formalized below.

\begin{restatable}{theorem}{inversebellman}
    \label{thm:inverse_bellman}
    Let $\CalT^{\pi}: \CalQ \rightarrow \CalR$ be the inverse Bellman operator (where $\CalQ$ and $\CalR$ are the spaces of value functions and reward functions, respectively) defined as:
    \begin{align}
        (\CalT^{\pi} \circ Q^{\pi})(\bss, \bsa) 
        = Q^{\pi}(\bss, \bsa) - \gamma \EE_{\subalign{\bss' &\sim T(\cdot | \bss, \bsa) \\ \bsa' &\sim \pi(\cdot | \bss')}} \big[ Q^{\pi}(\bss', \bsa') \big].
    \end{align}
    For a given transition $T$ in \cref{eq:mdp_transition} and policy $\pi$ in \cref{eq:boltzmann_policy}, $\CalT^{\pi}$ is a bijective mapping.
\end{restatable}
\begin{proof}[Sketch of proof] 
    We prove that the discretized Bellman operator is a linear operator represented by an invertible matrix in a vectorized representation of joint states $(\bss, \bsa)$.
    See Appendix \ref{appsec:inverse_bellman} or~\citet{garg2021iqlearn} for the complete proof. 
\end{proof}

This implies that estimating the potential function $\psi$ in the FP PDE 
corresponding to the induced MP 
is sufficient to recover the reward function in the MDP.

\subsection{Summary of the FP-IRL algorithm}
\label{sec:fpirl_algorithm}

The FP-IRL framework provides a physics-constrained approach to recovering reward functions and policies from observed behavior.
As outlined in \cref{alg:fpirl} and illustrated in \cref{fig:fpirl_summary}, 
the procedure begins by reformulating the original MDP as an MP over joint state-action variables. This reformulation enables a direct connection to the FP PDE described in \cref{eq:fokker_planck_eq}, providing a physics-informed representation of the distributional dynamics. 

To perform inference over the FP PDE, observed trajectory data is first converted into a density representation.
VSI is then applied to estimate the potential function $\psi$ that governs the system's drift dynamics.
Using the relationship established in \cref{eq:mdp_transition}, the corresponding MDP transition kernel $T$ is subsequently derived from this potential.

Leveraging \cref{thm:equivalence} and the principle of free energy minimization, the reward function $R$ and the optimal policy $\pi^{\ast}$ for the original MDP are recovered through closed-form expressions in \cref{eq:inverse_bellman,eq:boltzmann_policy}, respectively. 
Both quantities depend solely on the estimated potential function and can be computed efficiently with minimal overhead. 
This end-to-end approach provides a scalable, interpretable, and theoretically grounded method for IRL in continuous, stochastic, and physics-constrained environments.

\begin{algorithm}[htbp]
    \caption{Fokker--Planck IRL (FP-IRL)}
    \label{alg:fpirl}
    \KwIn{Observed trajectories $\CalD$; MDP with unknown reward and transition functions $\CalM \setminus \{ R, T \}$.}
    \KwOut{Estimated reward function $R(\bss,\bsa)$, policy $\pi(\bsa|\bss)$, and transition function $T(\bss'|\bss,\bsa)$.}
    Construct time-indexed state-action density $\{p_t\}$ from trajectories $\CalD$\;
    Infer potential function $\psi(\bsx)$ using VSI as described in \cref{sec:vsi}\;
    Recover transition function $T(\bss' | \bss, \bsa)$ using \cref{eq:gaussian_transition}\;
    Recover policy $\pi(\bsa | \bss)$ using the Boltzmann form in \cref{eq:boltzmann_policy}\;
    Recover reward function $R(\bss, \bsa)$ using the inverse Bellman equation in \cref{eq:inverse_bellman}.
\end{algorithm}

\section{Fokker--Planck PDE Inference via Variational System Identification}
\label{sec:vsi}
In this section, we discuss the use of VSI to infer the parameterized FP PDE. 
For detailed background on VSI, we refer readers to~\citet{Wang2019Variational,Wang2021Variational}. 

We consider the time-evolving probability density field $p(\bsx, t): \Omega \times [0, \tau] \to \RR^{+}$
where $\Omega = \prod_{i=1}^{d} \allowbreak [a_i, b_i]$
and $[0, \tau]$ is the time interval. 
For notational simplicity, we omit the explicit time dependence unless otherwise noted (writing $p_t(\bsx)$ when needed).
In the MDP context, $\bsx$ denotes the state-action pair,
serving as the analogue of spatial coordinates in statistical physics.
We focus on settings where both the density field 
$p(\bsx)$
and potential function $\psi(\bsx)$ are periodic in each spatial dimension. Specifically, for all $t \in [0, \tau]$, $\bsx \in \RR^d$, and $i\in \{1, \cdots, d\}$, we assume:
\begin{align}
    p \big(\bsx + (b_i - a_i) \bse_i \big) & =    p(\bsx), \\
    \nabla p \big(\bsx + (b_i - a_i) \bse_i \big) & = \nabla p(\bsx), \\
    \psi \big(\bsx + (b_i - a_i) \bse_i \big) & = \psi(\bsx), \\
    \nabla \psi \big(\bsx + (b_i - a_i) \bse_i \big) & = \nabla \psi (\bsx),
\end{align}
where $\bse_i$ is the unit vector in the $i$-{th} direction, and the Einstein summation convention holds. 

We pose the  FP PDE in its weak form with periodic boundary conditions, seeking solutions $p(\cdot) \in H^1_P(\Omega)$, where $H^1_P(\Omega)$ denotes the Sobolev space of square-integrable, periodic functions with square-integrable first derivatives. 
The weak form is obtained by multiplying \cref{eq:fokker_planck_eq} with weighting functions (i.e., test functions) $w(\bsx) \in H^1_P(\Omega)$, integrating over the domain, and applying the divergence theorem: 
\begin{align}
\begin{split}
    & \int_{\Omega} \pp{p}{t} w \, \mathrm{d}\Omega + \int_{\Omega} \left( p \nabla \psi \cdot \nabla w + \beta^{-1} \nabla p \cdot \nabla w \right) \, \mathrm{d}\Omega \\
    =& \int_{\partial \Omega} \left( w p \nabla \psi \cdot \bsn + \beta^{-1} w \nabla p \cdot \bsn \right) \, \mathrm{d}S,
\end{split}
\end{align}
where $\bsn$ denotes the outward unit normal of the domain boundary. 
Due to the periodicity boundary conditions, 
the boundary integral vanishes, and the weak form simplifies to:
\begin{align}
    \label{eq:fokker_planck_weak}
    \int_{\Omega} \pp{p}{t} w \, \mathrm{d}\Omega + \int_{\Omega} p\nabla \psi \cdot \nabla w + \beta^{-1} \nabla p \cdot \nabla w \, \mathrm{d} \Omega = 0.
\end{align}    
A function 
$p$ 
satisfying \cref{eq:fokker_planck_weak} for all weighting functions $w$ and prescribed initial condition $p_0$ is considered a weak solution to the FP PDE. 

The goal of VSI in this setting is to estimate both the potential function $\psi$ and the inverse temperature $\beta$ from empirical density data 
$ \{ p^{\text{data}}_{t_k} \}_{t_k=0}^{\tau}$
derived from observed trajectories $\CalD$. 
Because \cref{eq:gaussian_transition} for the transition function $T$ requires a differentiable potential, we seek a smooth approximation to $\psi$. 
In the following subsection, we describe a discretized representation of $\psi$ using a finite basis of differentiable functions over $\Omega$, enabling tractable numerical inference. 

\subsection{Hermite cubic interpolation for the potential function \texorpdfstring{$\psi$}{psi} }
\label{sec:hermite}

To satisfy the regularity requirements of the potential function $\psi$, we adopt a tensor-product basis of piecewise cubic \textit{Hermite polynomials} for interpolation.
The domain in each dimension $[a_i, b_i]$ is partitioned into $n_{h, i}$ non-overlapping elements as $[a_i, b_i] = \bigcup_{j=1}^{n_{h, i}} \big[x_i^{j}, x_i^{j+1} \big]$, with end points $x_i^1 = a_i$ and $x_i^{n_{h, i}+1} = b_i$, and $x_i^{j} < x_i^{j+1}$. 
For each such subinterval, we construct a one-dimensional Hermite basis $\CalB_i = \{h_1, \cdots, h_{2n_{h, i} + 2}\}$, consisting of standard cubic Hermite polynomials:
\begin{align*}
    \overline{h}_1(\xi) &= 1 - 3 \xi^2 + 2 \xi^3, \\
    \overline{h}_2(\xi) &= \xi - 2 \xi^2 + \xi^3, \\
    \overline{h}_3(\xi) &= 3 \xi^2 - 2 \xi^3, \\
    \overline{h}_4(\xi) &= -\xi^2 + \xi^3,
\end{align*}
which define the value and slope interpolation within each subinterval.
For any $j \in \{1, \cdots, n_{h, i}+1\}$, the one-dimensional Hermite basis functions centered at node $x_i^j$ are given by:
\begin{align}
    h_{2j-1}(x) &= 
        \begin{cases} 
            \overline{h}_1\left(\frac{x - x_i^j}{x_i^{j+1}-x_i^j }\right), & \text{if } x \in \big[x_i^j, x_i^{j+1}\big); \\
            \overline{h}_3\left(\frac{x - x_i^{j-1}}{x_i^j-x_i^{j-1}}\right), & \text{if } x \in \big[x_i^{j-1}, x_i^j \big) \text{ and } j \neq 1;  \\
            \overline{h}_3\left(\frac{x - x_i^{n_{h, i}}}{x_i^{n_{h, i} + 1}-x_i^{n_{h, i}}}\right), & \text{if } x \in \big[x_i^{n_{h, i}}, x_i^{n_{h, i}+1} \big) \text{ and } j = 1; \\
            0, & \text{otherwise};
        \end{cases} \\
    h_{2j}(x) &= 
        \begin{cases} 
            (x_i^{j+1}-x_i^j )\overline{h}_2\left(\frac{x - x_i^j}{x_i^{j+1}-x_i^j }\right), & \text{if } x \in \big[x_i^j, x_i^{j+1} \big); \\
            (x_i^j-x_i^{j-1})\overline{h}_4\left(\frac{x - x_i^{j-1}}{x_i^j-x_i^{j-1}}\right), & \text{if } x \in \big[x_i^{j-1}, x_i^j \big) \text{ and } j \neq 1;  \\
            (x_i^{n_{h, i} + 1}-x_i^{n_{h, i}})\overline{h}_4\left(\frac{x - x_i^{n_{h, i}}}{x_i^{n_{h, i} + 1}-x_i^{n_{h, i}}}\right), & \text{if } x \in \big[x_i^{n_{h, i}}, x_i^{n_{h, i}+1} \big) \text{ and } j = 1; \\
            0, & \text{otherwise}.
        \end{cases}
\end{align}
These basis functions allow any continuously differentiable, periodic function $f(x)$ over $[a_i, b_i]$ to be approximated as 
$$f(x) = \sum_{j=1}^{2n_{h, i} + 2} \theta_j h_{j}(x),$$
where $\theta_{2j-1}$ and $\theta_{2j}$ represent the function value and slope at node $x_i^j$, respectively. 
See~\citet[Chapter 4]{de1978practical} for further details on this interpolation scheme.

We extend this construction to the full $d$-dimensional domain by assembling the tensor product of the one-dimensional bases across all dimensions. The potential function $\psi(\bsx)$ is approximated as:
\begin{align}
    \label{eq:potential_funciton}
    \psi(\bsx) = \sum_{j_1,\cdots,j_d} \theta_{j_1,\cdots,j_d} \phi_{j_1,\cdots,j_d}(\bsx) , \quad \text{ with } j_i \in \{1, \cdots, 2n_{h, i} + 2\}, %
\end{align}
where each basis is defined as a tensor product:
\begin{align}
    \label{eq:hermite_basis}
    \phi_{j_1, \cdots, j_d} (\bsx)= h_{j_1}(x_1) \times \cdots \times h_{j_d}(x_d).
\end{align}

\subsection{Numerical discretization using finite element interpolation}
We construct a grid-based mesh over the $d$-dimensional hyper-rectangular domain $\Omega = \prod_{i=1}^{d} \allowbreak [a_i, b_i]$. 
Each dimension is divided into $n_{e, i}$ non-overlapping elements: $[a_i, b_i] = \bigcup_{j=1}^{n_{e, i}} \big[x_i^{j}, x_i^{j+1} \big]$, with $x_i^1 = a_i$, $x_i^{n_{e, i}+1} = b_i$, and $x_i^{j} < x_i^{j+1}$.
The resulting mesh comprises $n_e = \prod_{i=1}^{d} n_{e, i}$ elements and is generally chosen to be much finer than the mesh used to interpolate the potential function $\psi$.

Each 
element is constructed as a tensor product of grid nodes: $\Omega_{e=(j_1,\cdots, j_{d})} = \prod_{i=1}^{d} [x_i^{j_i}, x_i^{j_i+1}]$, where the grid nodes are $\big\{ \big(x_{1}^{j_1}, \cdots, x_{i}^{j_i}, \cdots, x_{d}^{j_d} \big) \big | \, i \in \{1, \cdots d\}, \allowbreak j_i \in \{1, \cdots, n_{e, i} + 1\} \big\}$. 
Within each element, we perform piecewise linear interpolation of the density field $p(\bsx)$ using standard finite element shape functions:
\begin{align}
\label{eq:interpolation_expanded}
    p(\bsx) = \sum_{l_1\in \{0,1\} }\cdots \sum_{l_d\in \{0,1\} }p(x_1^{j_1 +l_1}, \cdots, x_d^{j_d +l_d} ) \prod_{i=1}^{d} \overline{N}_{l_i} \left(\frac{x_i - x_i^{j_i}}{x_i^{j_i +1} - x_i^{j_i}}\right),
\end{align}
where the one-dimensional linear shape functions are:
\begin{align*}
    \overline{N}_{0}(\xi) &= 1 - \xi, \\
    \overline{N}_{1}(\xi) &= \xi.   
\end{align*}

This interpolation consists of $2^d$ basis terms corresponding to the corners of the hyper-rectangle. 
Each basis function is 1 at its corresponding node and 0 at all other nodes. 
For notational compactness, we write \cref{eq:interpolation_expanded} as:
\begin{align}
    p(\bsx) = \sum_{q=1}^{2^d} p_{e(q)} N_q(\bsx),
\end{align} 
where $N_q$ is the shape function for the $q$-{th} node of element $e$, and $p_{e(q)}$ is the associated nodal density value.

\subsection{Parameter estimation via residual minimization}
Given the potential function ansatz from \cref{eq:potential_funciton}, we now derive the residual form of the weak PDE to estimate the potential function coefficients and inverse temperature, collectively denoted by $\bstheta = \{\theta_{\bsi}\}_{\bsi} \cup \{ \beta \}$ where $\bsi = (i_1, \cdots, i_d)$ denotes the multi-index. The residual is given by:
\begin{align}
    \label{eq:fokker_planck_residual}
    \CalR = \int_{\Omega} \pp{p}{t} w \, \mathrm{d} \Omega 
    + \sum_{i_1,\cdots,i_d} \theta_{i_1,\cdots,i_d} \int_{\Omega} p \nabla \phi_{i_1,\cdots,i_d} \cdot \nabla w \, \mathrm{d} \Omega
    + \beta^{-1} \int_{\Omega}\nabla p \cdot \nabla w \, \mathrm{d} \Omega,
\end{align}    
where $p$ is interpolated from the values at the grid nodes. %
Following the Galerkin approach, 
we choose a set of weighting functions 
defined as:
\begin{align}
    w_{j_1, \cdots, j_d} = \prod_{i=1}^{d} 
    \begin{cases}
        \overline{N}_{0}\left(\frac{x_i - x_i^{j_i}}{x_i^{j_i +1} - x_i^{j_i}}\right), & \text{if } x_i \in \big[x_i^{j_i}, x_i^{j_i+1} \big); \\        
        \overline{N}_{1}\left(\frac{x_i - x_i^{j_i -1}}{x_i^{j_i } - x_i^{j_i - 1 }}\right), & \text{if } x_i \in \big[ x_i^{j_i-1}, x_i^{j_i} \big) \text{ and }  j_i \neq 1; \\ 
        \overline{N}_{1}\left(\frac{x_i - x_i^{n_{e, i}}}{x_i^{n_{e, i} +1} - x_i^{n_{e, i} }}\right), & \text{if } x_i \in \big[x_i^{n_{e, i}}, x_i^{n_{e, i} + 1} \big) \text{ and }  j_i = 1; \\
        0, & \text{otherwise}.
    \end{cases}
\end{align}
Evaluating the weak form for each weighting function yields a set of algebraic residual equations:
\begin{align}
    \label{eq:fokker_planck_residual_1}
    \CalR_{j_1, \cdots, j_d} =& \int_{\Omega} \pp{p}{t} w_{j_1, \cdots, j_d} \, \mathrm{d} \Omega
    + \sum_{i_1,\cdots,i_d} \theta_{i_1,\cdots,i_d} \int_{\Omega} p \nabla \phi_{i_1,\cdots,i_d} \cdot \nabla w_{j_1, \cdots, j_d} \, \mathrm{d} \Omega \nonumber\\
    &+ \beta^{-1} \int_{\Omega}\nabla p \cdot \nabla w_{j_1, \cdots, j_d} \, \mathrm{d} \Omega.
\end{align}
Collecting all such equations over all nodes gives a linear system in the unknowns $\btheta$. The residual system is written in matrix-vector form:
\begin{align}
    \boldsymbol{\CalR} = \bsy - [\bsXi_{\bsi}, \cdots, \bsXi_{\beta}] %
    [\theta_{\bsi}, \cdots, \beta^{-1}]^{\top},
\end{align}
where the $\bsj$-{th} entry of the vectors $\boldsymbol{y}$ and the columns of matrix $\bsXi$ are evaluated as: 
\begin{align}
    \label{eq:fokker_planck_residual_y}
    y_{\bsj} &= \sum_{e} \sum_{q=1}^{2^d} \int_{\Omega_e} \pp{p_{e(q)}}{t} N_{e(q)}w_{\bsj} \, \mathrm{d} \Omega, \\
    \Xi_{\bsi, \bsj} &= \sum_e \sum_{q=1}^{2^d} \int_{\Omega_e} p_{e(q)} N_{e(q)} \nabla \phi_{\bsi}\cdot \nabla w_{\bsj} \, \mathrm{d} \Omega, \\
    \Xi_{\beta, \bsj} &= \sum_e \sum_{q=1}^{2^d} \int_{\Omega_e} p_{e(q)} \nabla N_{e(q)} \cdot \nabla w_{\bsj} \, \mathrm{d} \Omega.
\end{align}
These integrals are evaluated numerically using Gaussian quadrature. 

The solution $p(\bsx)$ for known coefficients $\boldsymbol{\theta}$ would yield a zero residual for all weighting functions and at all time steps.
In practice, we solve the following least squares problem for unknown $\boldsymbol{\theta}$: 
\begin{align}
    \label{eq:vsi_regression}
    \boldsymbol{\theta}^{\ast} = \argmin_{\bstheta} \sum_{t \in [0, \tau]}
    \norm{\boldsymbol{\CalR} \( p^{\text{data}}(\cdot, t) ; \bstheta \)}{2}^2,
\end{align}
where the residual $\boldsymbol{\CalR}$ is evaluated using the observed density field $p^{\text{data}}(\bsx,t)$ at discrete time steps $t \in [0, \tau]$. 
A sufficiently small residual indicates that the estimated parameters define an FP PDE consistent with the observed dynamics.

\subsection{Uniqueness of the potential (or value) function}
\label{sec:value_function_bias}

The parameterization of the potential function $\psi$ in \cref{eq:potential_funciton} intentionally omits the constant term. This is because both the residual formulation in \cref{eq:fokker_planck_residual} and the least-squares estimation in \cref{eq:vsi_regression} depend only on the gradient of $\psi$. As a result, the recovered potential function is determined only up to an additive constant.

This inherent ambiguity has no impact on the inferred system dynamics or policy.
Let $\hpsi$ be the estimated potential, such that $\hat{\psi}(\bss, \bsa) = \psi(\bss, \bsa) + c$ for some constant $c$ and all  $ (\bss,\bsa) \in \CalS \times \CalA$.  
By \cref{thm:equivalence}, the value function is given by ${Q}^{\pi}= -\psi$, so the estimated value function becomes $\hat{Q}^{\pi}=-\hat{\psi} = Q^{\pi}- c$, also differing from the true value by the same constant.

The transition function $T(\bss'|\bss,\bsa)$, which depends only on the gradient $\nabla \psi$ as shown in \cref{eq:gaussian_transition}, is invariant under constant shifts in $\psi$.
Similarly, the policy defined via the Boltzmann distribution in \cref{eq:boltzmann_policy} is unaffected by such shifts. Explicitly, using the estimated value function $\hat{Q}^{\pi}$, the policy becomes:
\begin{align}
    \pi(\bsa|\bss) = 
    \frac{\exp \big(\beta \hat{Q}^{\pi}(\bss, \bsa) \big)}{\int_{\CalA} \exp \big(\beta \hat{Q}^{\pi}(\bss, \bsa') \big) \,\mathrm{d} \bsa'} 
    =& \frac{\exp \big(\beta Q^{\pi}(\bss, \bsa) + \beta c \big)}{\int_{\CalA} \exp \big(\beta Q^{\pi}(\bss, \bsa') + \beta c \big) \,\mathrm{d} \bsa'} \nonumber\\
    =& \frac{\exp(\beta c) \exp \big(\beta Q^{\pi}(\bss, \bsa) \big)}{\exp(\beta c) \int_{\CalA} \exp \big(\beta Q^{\pi}(\bss, \bsa') \big) \,\mathrm{d} \bsa'} \nonumber\\
    =& \frac{\exp \big(\beta Q^{\pi}(\bss, \bsa) \big)}{\int_{\CalA} \exp \big(\beta Q^{\pi}(\bss, \bsa') \big) \,\mathrm{d} \bsa'}.
\end{align}
Thus, the policy $\pi(\bsa|\bss)$ remains unchanged.

Moreover, the inverse Bellman equation in \cref{eq:inverse_bellman} relies exclusively on the transition dynamics, policy, and value function. As a result, a constant shift in the value function induces the same shift in the recovered reward function, reflecting the well-known fact that adding a constant to the objective does not change the optimal solution.

\section{Numerical Experiments}
\label{sec:experiments}
We begin by demonstrating the effectiveness of FP-IRL on a controlled synthetic example based on the classical Grid World problem.
Standard RL benchmarks, such as those found in OpenAI Gym, are not directly applicable, as their state-action dynamics generally do not adhere to the FP formulation required by our method. To highlight the broader applicability of FP-IRL, we also include a modified version of the well-known Mountain Car problem, adapted to satisfy FP dynamics.

\Cref{tab:comutational_complexity} summarizes the computational complexity across experiments.
Memory requirements scale as $\CalO(n^{d})$, where $n$ is the number of discretization nodes per dimension, and $d$ is the problem dimensionality.
Due to this exponential scaling, memory usage becomes a bottleneck, and so we restrict our experiments to $d\leq 4$.

\subsection{Synthetic Grid World example}
\label{sec:synthetic_example}

\begin{figure}[htbp]
    \centering
    \includegraphics[width=\textwidth]{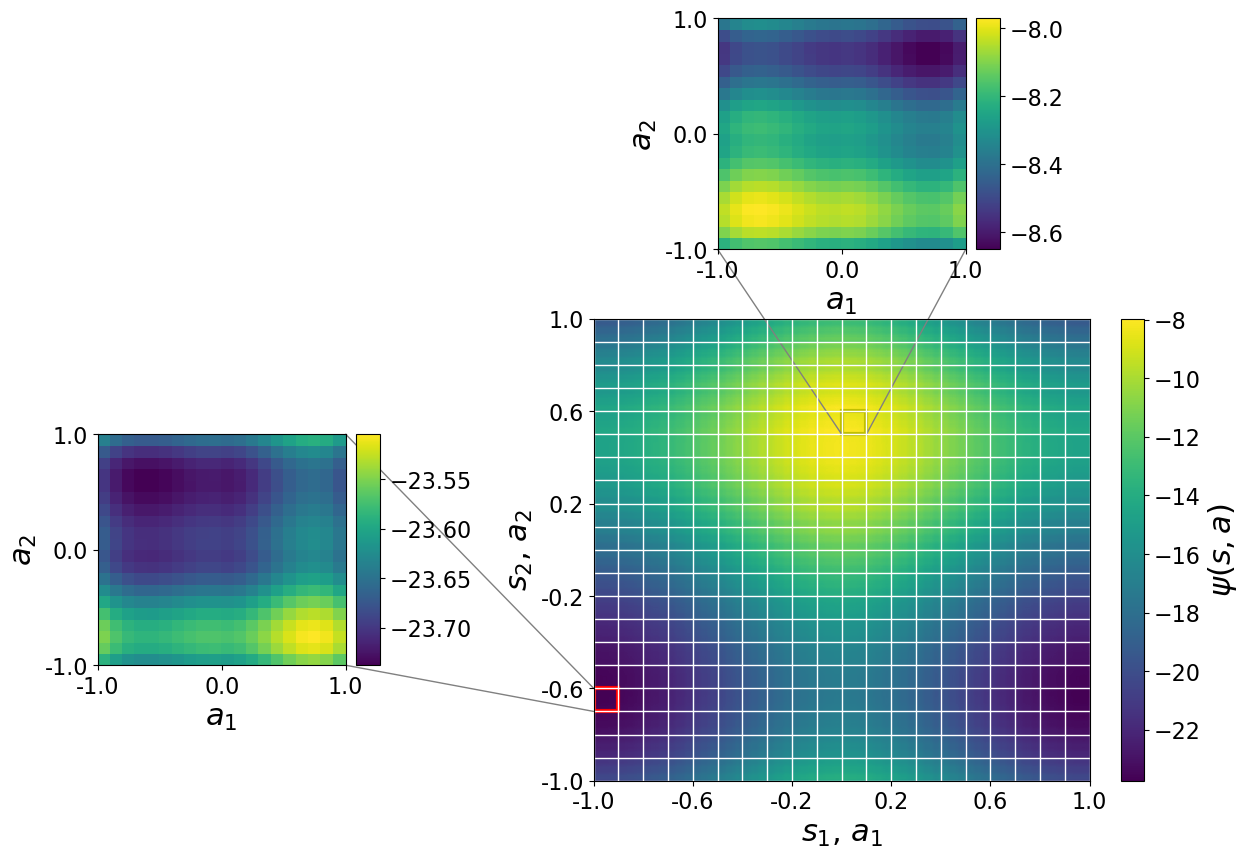}
    \caption{Grid World case. Visualization of the prescribed ground-truth potential function $\psi_{\text{GT}}$, defined over a four-dimensional state-action space $\CalS \times \CalA$. The function is shown on a
    $20 \times 20$ grid, where the primary grid axes correspond to the state variables $s_1$ and $s_2$, and each cell contains a sub-grid representing the variation over action variables $a_1$ and $a_2$.
    High potential values are concentrated near the top center of the domain, while lower values are located in the bottom corners.
    The color scale encodes the potential function value.
    Two representative sub-grids 
    are highlighted: one at the top (high-potential regions) and one on the left (low-potential regions), illustrating the local structure of the potential over actions at fixed states.
    }
    \label{fig:gridworld_value_gt_zoomin}
\end{figure}

To validate FP-IRL against known ground truth and assess convergence behavior, we construct a synthetic MDP set in a four-dimensional Grid World environment. The system dynamics are governed by the FP PDE (\cref{eq:fokker_planck_eq}), with the transition function adhering to \cref{eq:mdp_transition,eq:gaussian_transition}.  
We explicitly prescribe a ground-truth potential function $\psi_{\text{GT}}$ 
over the domain $[-1, 1]^4$, constructed
using the Hermite polynomial basis (\cref{eq:hermite_basis}) to ensure sufficient expressivity. 
The parameters used to define $\psi_{\text{GT}}$ are available in our code repository, and a visualization is provided in \cref{fig:gridworld_value_gt_zoomin}.

\begin{figure}[htbp]
    \centering
    \begin{subfigure}[t]{0.49\textwidth}
        \centering
        \includegraphics[height=4.4cm]{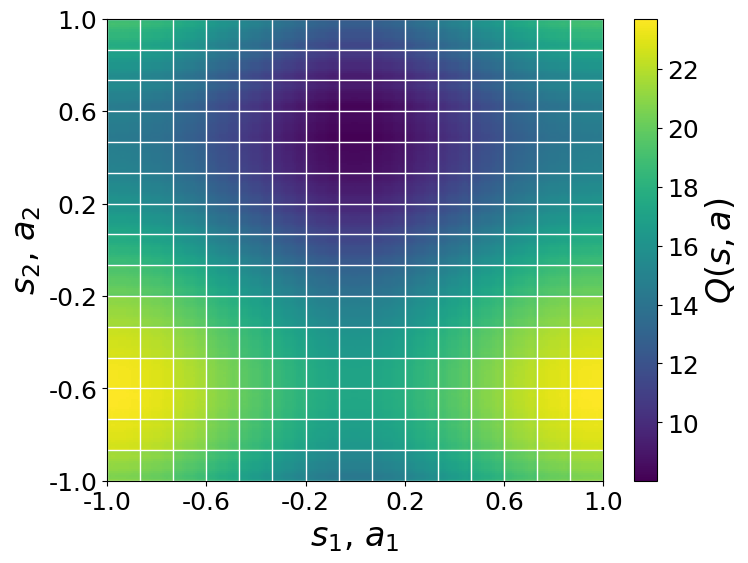}
        \caption{Ground-truth Q-function}
        \label{fig:gridworld_value_gt}
    \end{subfigure}
    \hfill
    \begin{subfigure}[t]{0.49\textwidth}
        \centering
        \includegraphics[height=4.4cm]{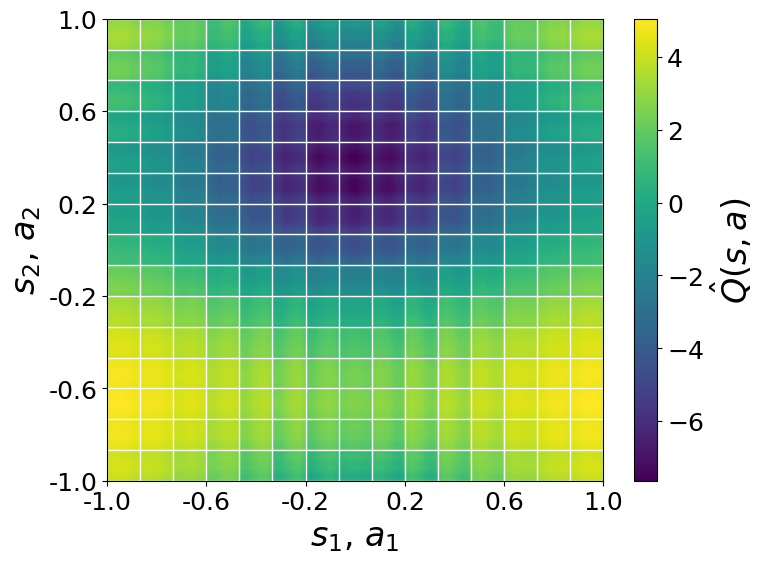}
        \caption{FP-IRL inferred Q-function}
        \label{fig:gridworld_value_vsi}
    \end{subfigure} 
    \begin{subfigure}[t]{0.49\textwidth}
        \centering
        \includegraphics[height=4.4cm]{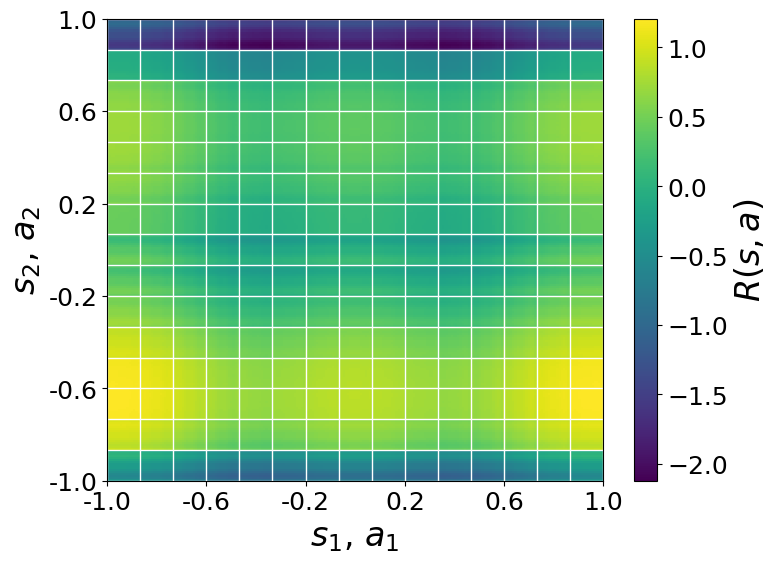}
        \caption{Ground-truth reward}
        \label{fig:gridworld_reward_gt}
    \end{subfigure}
    \hfill
    \begin{subfigure}[t]{0.49\textwidth}
        \centering
        \includegraphics[height=4.4cm]{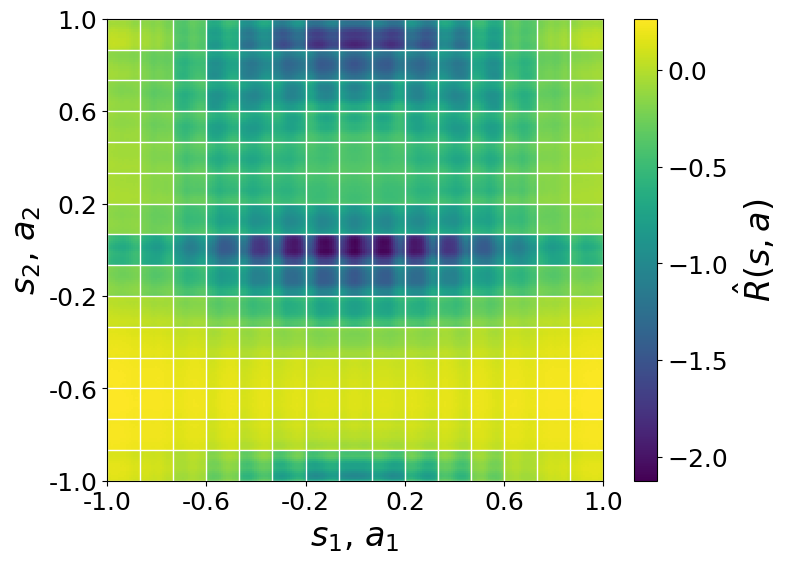}
        \caption{FP-IRL inferred reward}
        \label{fig:gridworld_reward_vsi}
    \end{subfigure}
    \begin{subfigure}[t]{0.49\textwidth}
        \centering
        \includegraphics[height=4.4cm]{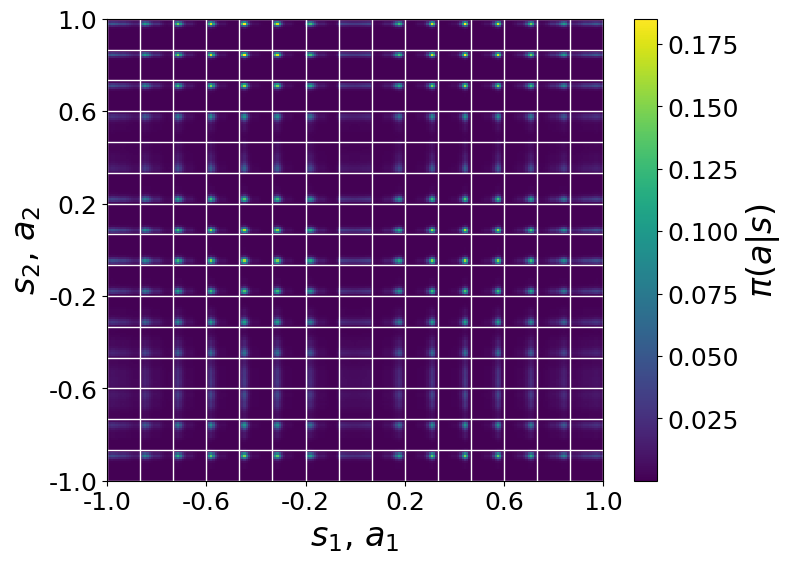}
        \caption{Ground-truth policy}
        \label{fig:gridworld_policy_gt}
    \end{subfigure}
    \hfill
    \begin{subfigure}[t]{0.49\textwidth}
        \centering
        \includegraphics[height=4.4cm]{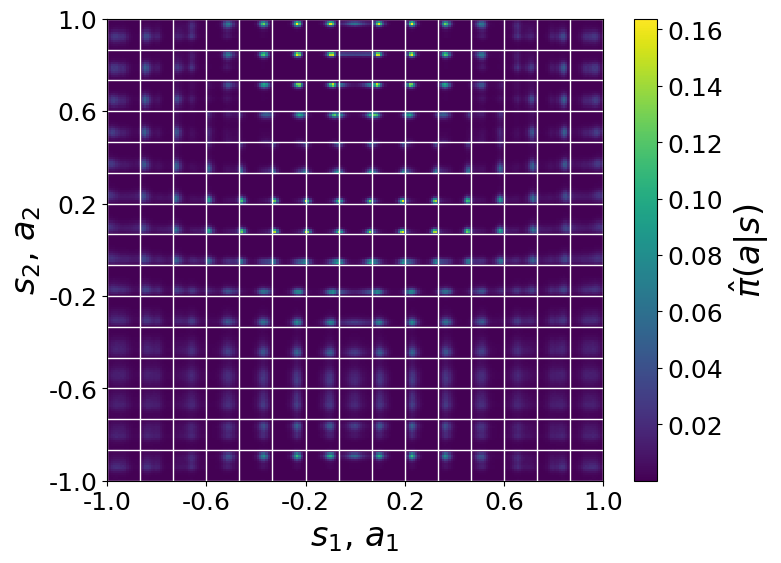}
        \caption{FP-IRL inferred policy}
        \label{fig:gridworld_policy_vsi}
    \end{subfigure}
    \caption{Grid World case. Comparison of ground-truth and inferred functions, computed on the highest-resolution mesh with partition size $N = 15$.
    Each panel displays one of the key functions: Q-function, reward, or policy, with left panels (a), (c), (e) showing the ground truth and right panels (b), (d), (f) showing the inferred counterparts. 
    The functions over $(s_1,s_2,a_1,a_2)$ are visualized using outer grids indexed by state variables $(s_1,s_2)$, and inner sub-grids for action variables $(a_1,a_2)$. Color represents the function value at each point in $\CalS\times\CalA$.
    Note that the Q-function is only determined up to an additive constant (cf. \cref{sec:value_function_bias}), so visual discrepancies between (a) and (b) are expected and do not affect the correctness of the inferred policy or reward.
    }
    \label{fig:gridworld_value_reward_policy} 
\end{figure}

From the prescribed potential, we obtain the ground-truth Q-function as $Q = -\psi_{\text{GT}}$, as established in \cref{thm:equivalence}.
The corresponding transition function $T$ and reward function $R$ are then computed using \cref{eq:mdp_transition,eq:inverse_bellman}, respectively. 
The optimal policy $\pi^{\ast}$ is derived via \cref{eq:boltzmann_policy,eq:conjecture_equivalence}, based on $\psi_{\text{GT}}$. 
The resulting ground-truth Q-function, reward, and policy are shown in \cref{fig:gridworld_value_gt,fig:gridworld_reward_gt,fig:gridworld_policy_gt}.

\begin{figure}[htbp]
    \centering
    \begin{subfigure}[t]{0.3\textwidth}
        \centering
        \includegraphics[height=3.5cm, trim={0 0 5cm 0}, clip]{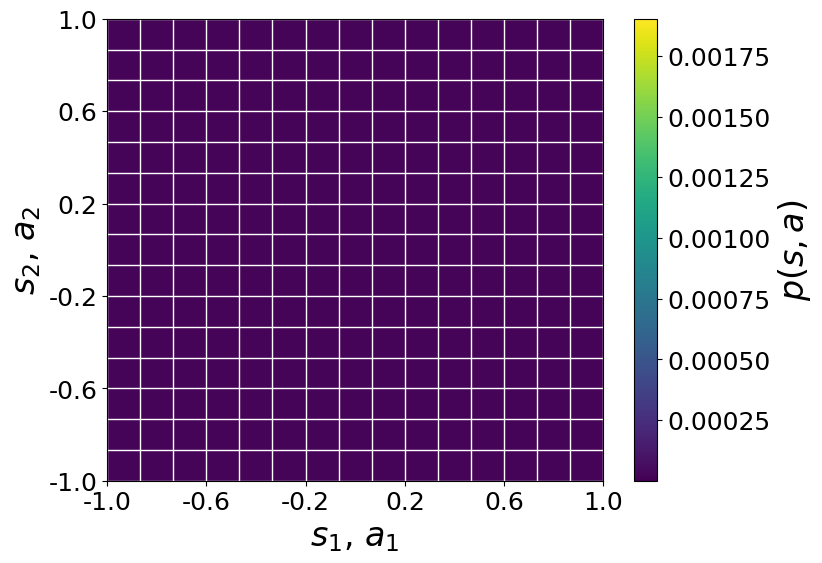}
        \caption{$t = 0$}
    \end{subfigure}
    \hfill
    \begin{subfigure}[t]{0.3\textwidth}
        \centering
        \includegraphics[height=3.5cm, trim={0 0 5cm 0}, clip]{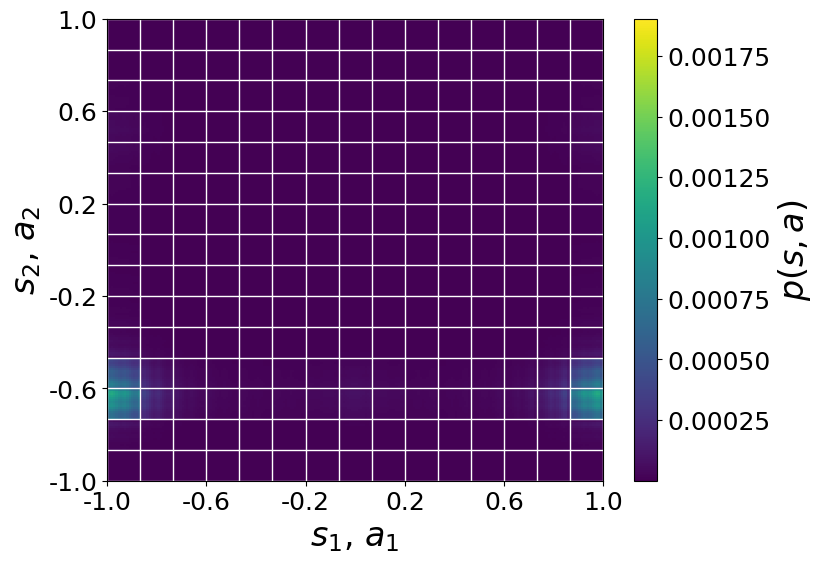}
        \caption{$t = 25$}
        \label{fig:gridworld_psa_gt_25}
    \end{subfigure}
    \hfill
    \begin{subfigure}[t]{0.38\textwidth}
        \centering
        \includegraphics[height=3.5cm]{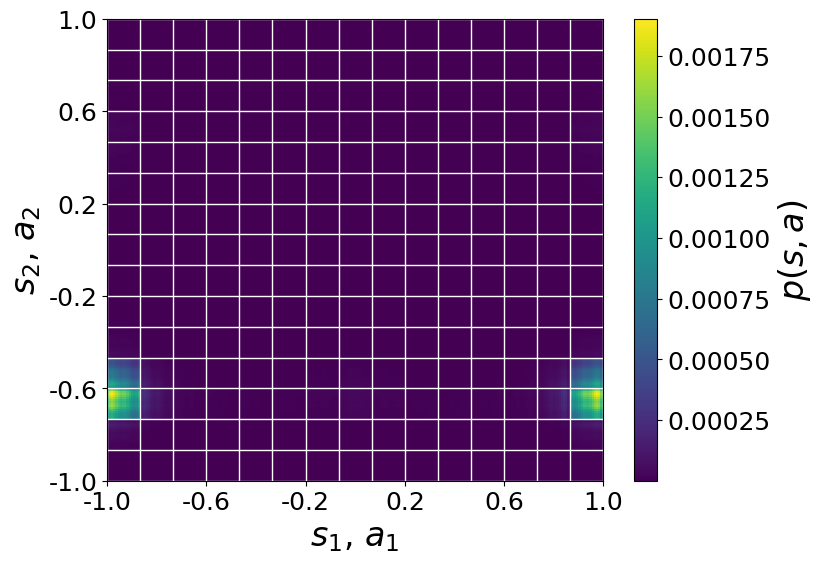}
        \caption{$t = 50$}
        \label{fig:gridworld_psa_gt_50}
    \end{subfigure}
    \begin{subfigure}[t]{0.3\textwidth}
        \centering
        \includegraphics[height=3.5cm, trim={0 0 4.5cm 0}, clip]{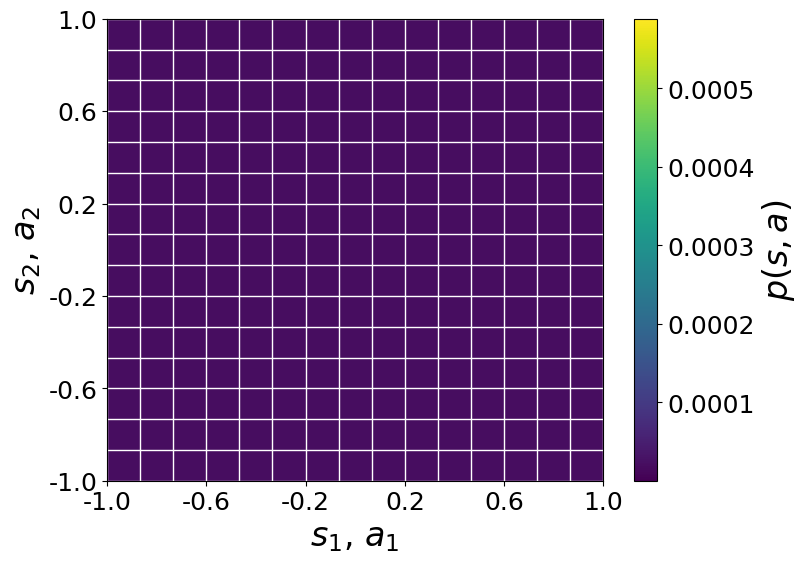}
        \caption{$t = 0$}
    \end{subfigure}
    \hfill
    \begin{subfigure}[t]{0.3\textwidth}
        \centering
        \includegraphics[height=3.5cm, trim={0 0 4.5cm 0}, clip]{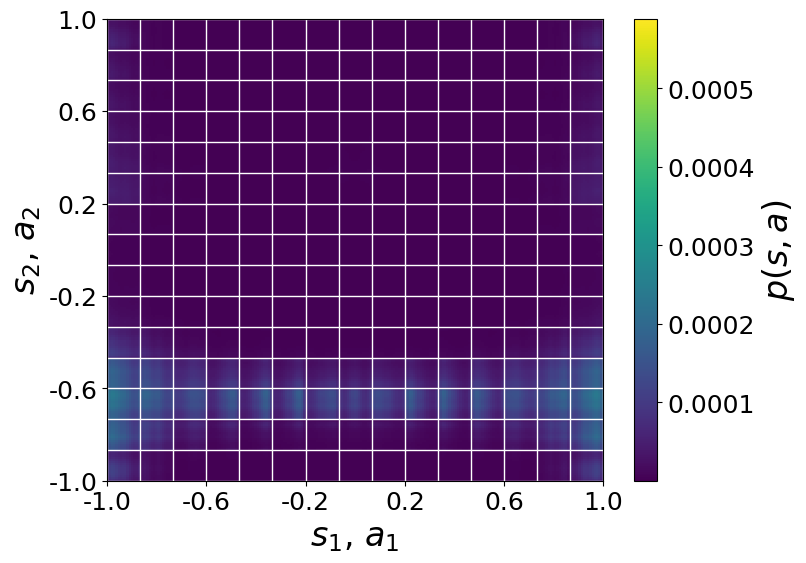}
        \caption{$t = 25$}
        \label{fig:gridworld_psa_vsi_25}
    \end{subfigure}
    \hfill
    \begin{subfigure}[t]{0.38\textwidth}
        \centering
        \includegraphics[height=3.5cm]{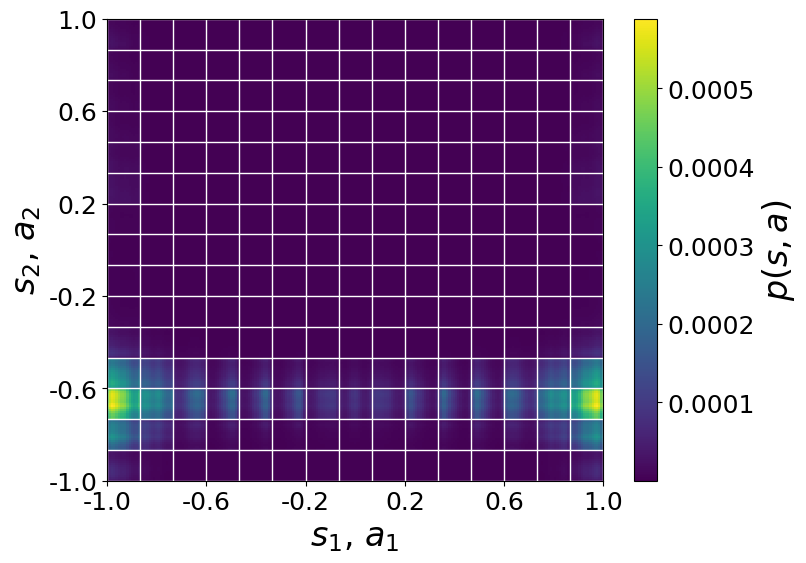}
        \caption{$t = 50$}
        \label{fig:gridworld_psa_vsi_50}
    \end{subfigure}
    \caption{Grid World case. Joint probability density $p_t$ of state-action pairs over time, computed on a mesh with partition size $N=15$. The top panels (a)--(c) depict the ground-truth probability densities at selected time steps, while the bottom panels (d)--(f) show the corresponding inferred probability densities obtained using FP-IRL. Each panel represents the four-dimensional state-action space using primary grid indexed by the state variables $(s_1,s_2)$, with embedded sub-grids capturing variations over action variables $(a_1,a_2)$. Color intensity indicates the density magnitude over $\CalS \times \CalA$. }
    \label{fig:gridworld_psa}
\end{figure}

To generate the observed data in the form of time-evolving densities $\CalD_p = \big\{ p_{t_k}^{\text{data}} \big\}_{t_k=0}^{\tau}$, 
we initialize the system with a uniform distribution 
$\rho_0 (\bss) = 1 / |\mathcal{S}|$
and compute the density evolution directly from the prescribed transition and policy:
\begin{align}
    \label{eq:density_evolution}
    p_{t_{n}}(\bss', \bsa') &= \pi^{\ast}(\bsa' | \bss') \int_{\CalS} \int_{\CalA} {p_{t_{n-1}} (\bss, \bsa)}  T(\bss' | \bss, \bsa)\, \mathrm{d} \bsa \,\mathrm{d} \bss.
\end{align}
The resulting probability density evolution is shown in \cref{fig:gridworld_psa}.
Alternately, individual trajectories 
$\big\{ \{ (\bss_{t_k}^{(i)}, \allowbreak \bsa_{t_k}^{(i)}) \}_{t_k=0}^{\tau} \big\}_{i=1}^m$ 
can be generated by Monte Carlo sampling from the transition and policy, with probability densities estimated using techniques such as kernel density estimation. 

Using the synthetic dataset 
$\CalD_p$, 
we apply FP-IRL in \cref{alg:fpirl} to infer the transition and reward functions and recover the optimal policy.
We begin by estimating the potential function $\psi$ 
via the VSI method described in \cref{sec:vsi}; 
in this example, we treat $\beta$ fixed to help simplify the problem.
The transition function is reconstructed by substituting the inferred potential into \cref{eq:mdp_transition}.
The inferred Q-function 
(via \cref{thm:equivalence}),
reward 
(via \cref{eq:inverse_bellman}),
and policy 
(via \cref{eq:boltzmann_policy})
are visualized in \cref{fig:gridworld_value_vsi,fig:gridworld_reward_vsi,fig:gridworld_policy_vsi}.
Comparison between the inferred and ground-truth functions demonstrates accurate recovery when using a high-resolution discretization of the state-action space. Some discrepancies remain, likely due to limitations in mesh resolution. This issue is explored further in a convergence study below.

\begin{figure}[htbp]
    \centering
    \begin{subfigure}{0.49\textwidth}
        \centering
        \includegraphics[height=4.5cm]{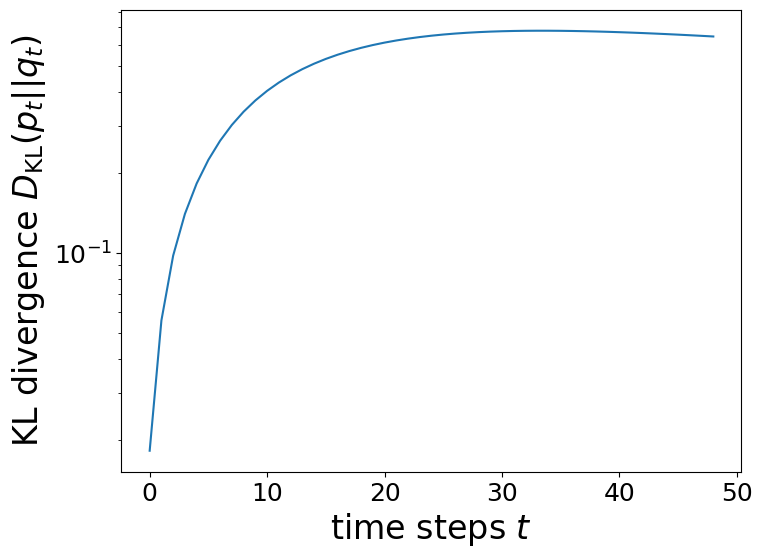}
        \caption{Full time interval ($t \in [0, 50]$).}
    \end{subfigure}
    \hfill
    \begin{subfigure}{0.49\textwidth}
        \centering
        \includegraphics[height=4.5cm]{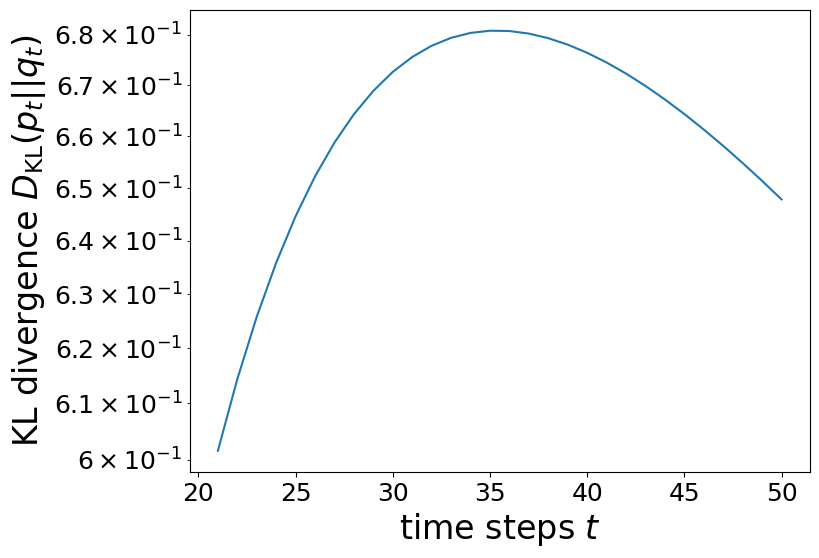}
        \caption{Zoomed-in interval ($t \in [20, 50]$).}
    \end{subfigure}
    \caption{Grid World case. KL divergence $\DKL(p_t^{\text{data}} || q_t)$ between the reference probability distribution $p_t^{\text{data}}$ from the data and the simulated distribution $q_t$, generated using the inferred policy and transition function. The divergence quantifies how closely the inferred dynamics match the observed data over time, with increasing divergence potentially reflecting accumulated inference errors.
    }
    \label{fig:kl_psa}
\end{figure}

To assess the fidelity of the inferred dynamics, we compare the simulated joint density generated from the recovered transition and policy, against the ground truth (\cref{fig:gridworld_psa}). 
The results show strong agreement.  
We also compute the {Kullback--Leibler} (KL) divergence, $\DKL(p_t^{\text{data}} || q_t)$, between the 
observed distribution $p_t^{\text{data}} \in \CalD_p$ 
and the simulated distribution $q_t$ generated using the inferred model (\cref{fig:kl_psa}).
The KL divergence increases modestly over time, likely due to accumulation of errors in the inferred dynamics (see \cref{fig:gridworld_psa}).
However, as shown in \cref{fig:kl_psa}, while the KL divergence is growing modestly when $t \leq 35$, it is decreasing  as $t \to 50$. 
These trends further light on the relative errors in the joint probability density $p_t$ at $t = 25$ by comparing \cref{fig:gridworld_psa_gt_25} versus \cref{fig:gridworld_psa_vsi_25}, and $t = 50$ by comparing \cref{fig:gridworld_psa_gt_50} versus \cref{fig:gridworld_psa_vsi_50}.

\begin{figure}[htbp]
    \centering
    \begin{subfigure}[t]{0.49\textwidth}
    \centering
        \includegraphics[height=4.5cm]{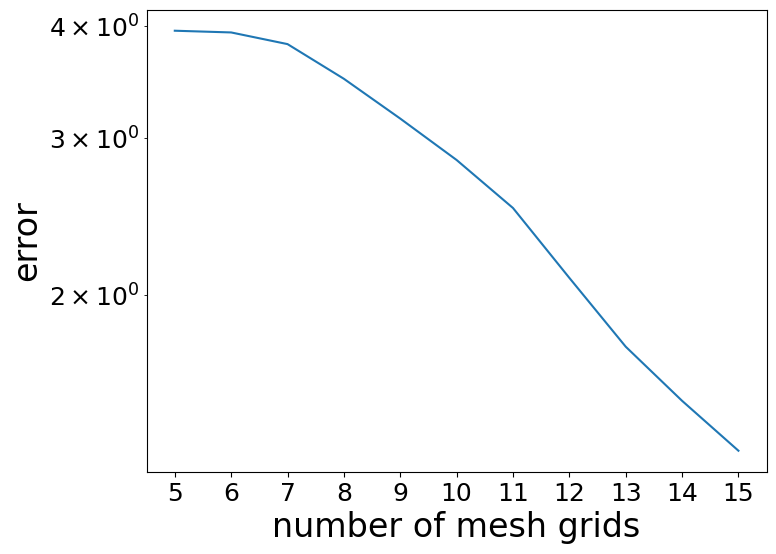}
        \caption{Error of potential function}
        \label{fig:gridworld_psi_error}
    \end{subfigure}
    \hfill
    \begin{subfigure}[t]{0.49\textwidth}
    \centering
        \includegraphics[height=4.5cm]{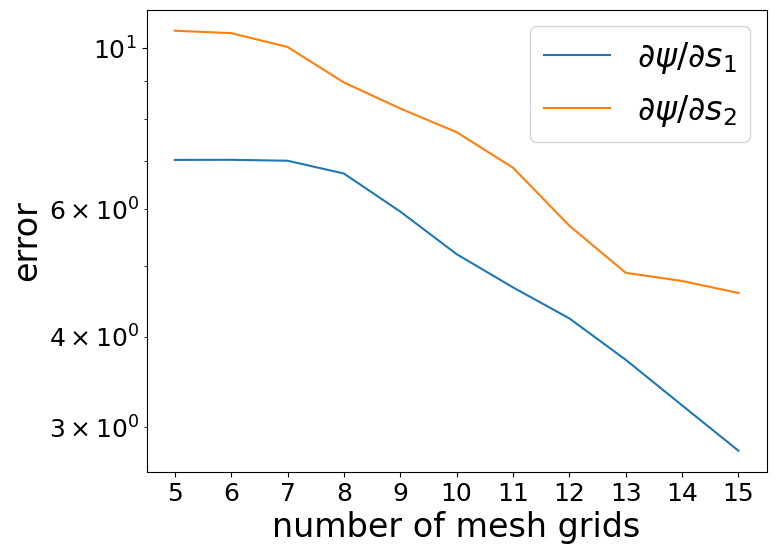}
        \caption{Error of the derivatives of potential function}
        \label{fig:gridworld_dpsi_error}
    \end{subfigure}
    \caption{Grid World case. Convergence analysis of the inferred value function and its derivatives with respect to mesh resolution. The plots show how the errors in the estimated value function $\hat{Q}(\bss, \bsa) = -\hat{\psi}(\bss, \bsa)$ and its spatial derivatives $\partial \hat{Q} / \partial \bss$ decrease as the number of partitions $N$ in each dimension increases. This demonstrates the expected convergence behavior of the FP-IRL framework under mesh refinement.}
    \label{fig:gridworld_convergence}
\end{figure}

Finally, we investigate convergence with respect to mesh resolution of the joint state-action space $\Omega$. 
Previous studies~\cite{Wang2019Variational,Wang2021Variational} have shown convergence for VSI. 
Here, we examine the convergence of the estimated potential function $\hat{\psi}$ and its derivatives $\pp{\hat{\psi}}{\bss}$, which directly affect the transition function via \cref{eq:mdp_transition}.
We consider uniform Cartesian meshes over the hypercube  $\Omega = [-1,1]^4$ with node locations 
$\bsx\in \lbrace -1, -1+\frac{2}{N}, \cdots ,  -1+\frac{2i}{N},\cdots, 1 \rbrace^4$ and vary the resolution from 5 to 15 nodes per dimension. 
We compute the error between the estimated and ground-truth functions using the $L^2$ norm: 
\begin{align}
    \text{error}(f) &= \left(\frac{1}{|\Omega|} \int_{\Omega} (f(\bsx) - f_{\text{GT}}(\bsx))^2 \,\mathrm{d}\bsx\right)^{\frac{1}{2}},
\end{align}
for $\Omega = \mathcal{S}\times\mathcal{A}$. The resulting error trends, shown in \cref{fig:gridworld_convergence}, confirm that both the potential function and its derivatives converge as resolution increases.

\subsection{Modified Mountain Car example}
\label{sec:openai_moutain_car}

Standard RL benchmarks, such as those provided in OpenAI Gym, are not directly compatible with FP-IRL, as their state-action dynamics do not naturally conform to the FP framework.
In this section, we demonstrate how to adapt a problem to fit with FP dynamics and apply our method to a modified version of the Mount Car example. 

The Mountain Car problem~\cite{Moore90efficientmemory} is a classic RL task where an underpowered vehicle must climb up a steep hill by building momentum. The state is defined by the car's position $x \in [-1.2, 0.6]$ and velocity $v \in [-0.07, 0.07]$, while the action is the applied force $a \in [-1, 1]$.

\begin{figure}[htbp]
    \centering
    \begin{subfigure}[t]{0.49\textwidth}
        \centering
        \includegraphics[height=4.4cm]{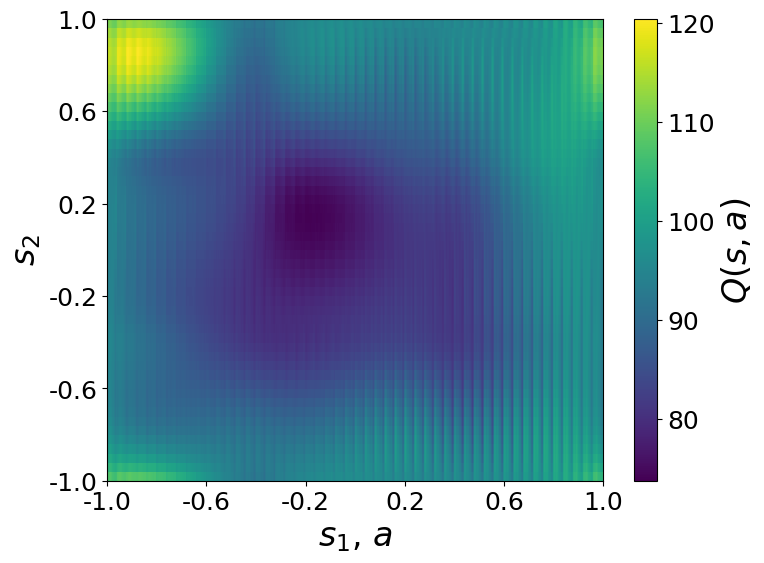}
        \caption{Ground-truth Q-function}
        \label{fig:moutaincar_q_value_gt}
    \end{subfigure}
    \hfill
    \begin{subfigure}[t]{0.49\textwidth}
        \centering
        \includegraphics[height=4.4cm]{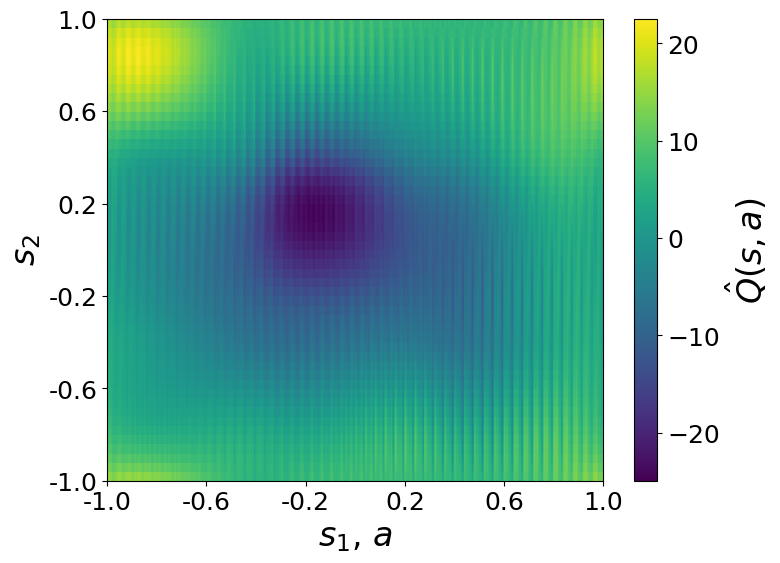}
        \caption{FP-IRL inferred Q-function}
        \label{fig:moutaincar_q_value_vsi}
    \end{subfigure}
    \hfill
    \begin{subfigure}[t]{0.49\textwidth}
        \centering
        \includegraphics[height=4.4cm]{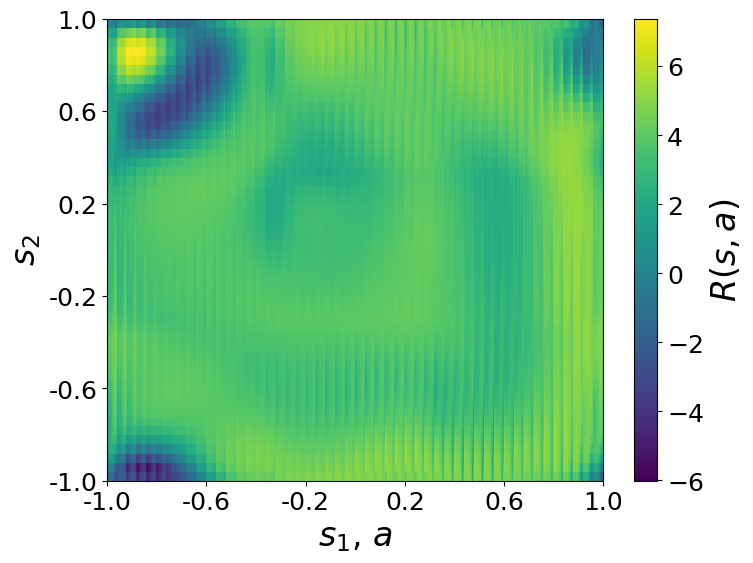}
        \caption{Ground-truth reward}
        \label{fig:mountain_car_reward_gt}
    \end{subfigure}
    \hfill
    \begin{subfigure}[t]{0.49\textwidth}
        \centering
        \includegraphics[height=4.4cm]{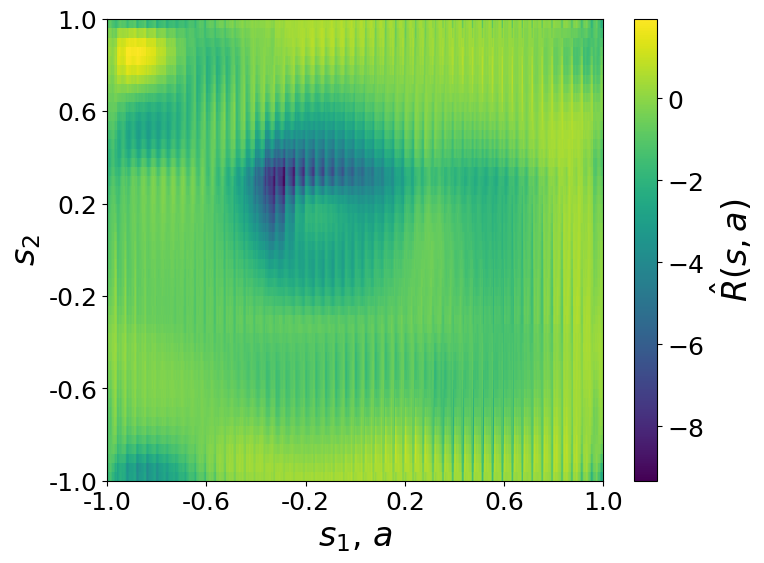}
        \caption{FP-IRL inferred reward}
        \label{fig:moutaincar_reward_vsi}
    \end{subfigure}
    \hfill
    \begin{subfigure}[t]{0.49\textwidth}
        \centering
        \includegraphics[height=4.4cm]{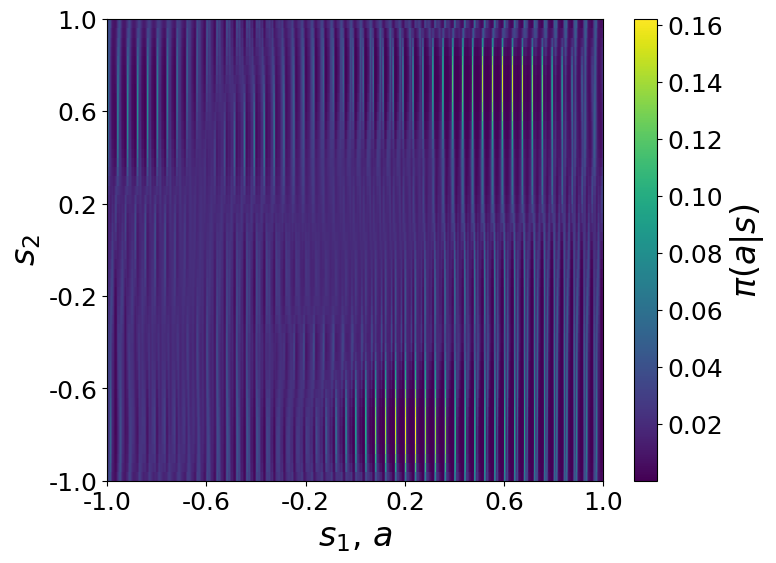}
        \caption{Ground-truth policy}
        \label{fig:mountain_car_policy_gt}
    \end{subfigure}
    \hfill
    \begin{subfigure}[t]{0.49\textwidth}
        \centering
        \includegraphics[height=4.4cm]{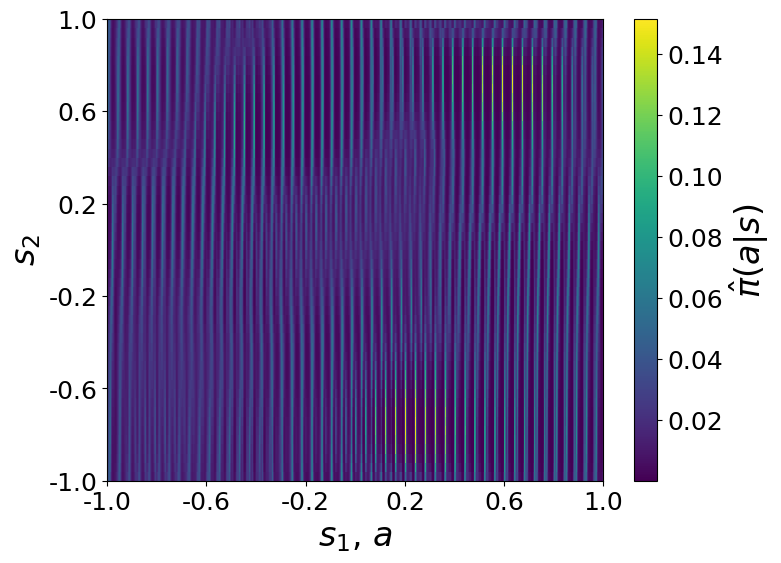}
        \caption{FP-IRL inferred policy}
        \label{fig:moutaincar_policy_vsi}
    \end{subfigure}
    \caption{Mountain Car case. Comparison of ground-truth and inferred functions, computed on the highest-resolution mesh with partition size $N = 50$.
    Each panel displays one of the key functions: Q-function, reward, or policy, with left panels (a), (c), (e) showing the ground truth and right panels (b), (d), (f) showing the inferred counterparts. 
    The functions over $(s_1,s_2,a)$ are visualized using outer grids indexed by state variables $(s_1,s_2)$, and inner sub-grids for action variables $a$. Color represents the function value at each point in $\CalS\times\CalA$.
    Note that the Q-function is only determined up to an additive constant (cf. \cref{sec:value_function_bias}), so visual discrepancies between (a) and (b) are expected and do not affect the correctness of the inferred policy or reward.
    }
    \label{fig:moutaincar_value_reward_policy}
\end{figure}

\paragraph{Adapting Mountain Car to FP dynamics}
The original Mountain Car system is governed by deterministic ODEs, making it unsuitable for direct use in the FP framework, which describes stochastic systems via PDEs.
To transform this setup, we proceed as follows:
\begin{enumerate}
    \item We first solve the original Mountain Car problem using a standard RL algorithm (e.g., soft actor-critic~\cite{haarnoja2018soft}) to
    obtain an approximate optimal Q-function. 

    \item We normalize the state-action space to the range of $[-1, 1]^{3}$ and construct Hermite basis functions (as in \cref{eq:hermite_basis}) on the normalized domain. 
    We then fit the learned Q-function onto the Hermit basis by minimizing the mean squared error.

    \item The interpolated Q-function (i.e., the negative FP potential function in \cref{eq:fokker_planck_eq}) defines the modified Mountain Car problem. \Cref{fig:moutaincar_q_value_gt} shows the resulting Hermite-based potential function. 

    \item We compute the time-evolving probability densities $\CalD_p = \big\{ p_{t_k}^{\text{data}} \big\}_{t_k=0}^{\tau}$ using the FP-compatible transition function and policy, following the procedures described in \cref{sec:synthetic_example} and \cref{eq:density_evolution}. 
\end{enumerate}

\begin{figure}[htbp]
    \centering
    \begin{subfigure}[t]{0.3\textwidth}
        \centering
        \includegraphics[height=3.6cm, trim={0 0 4.5cm 0}, clip]{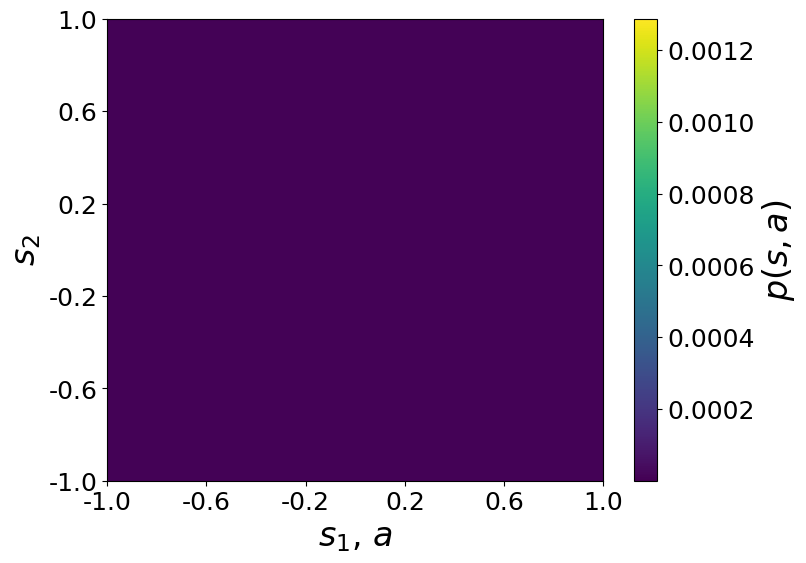}
        \caption{$t=0$}
    \end{subfigure}
    \hfill
    \begin{subfigure}[t]{0.3\textwidth}
        \centering
        \includegraphics[height=3.6cm, trim={0 0 4.5cm 0}, clip]{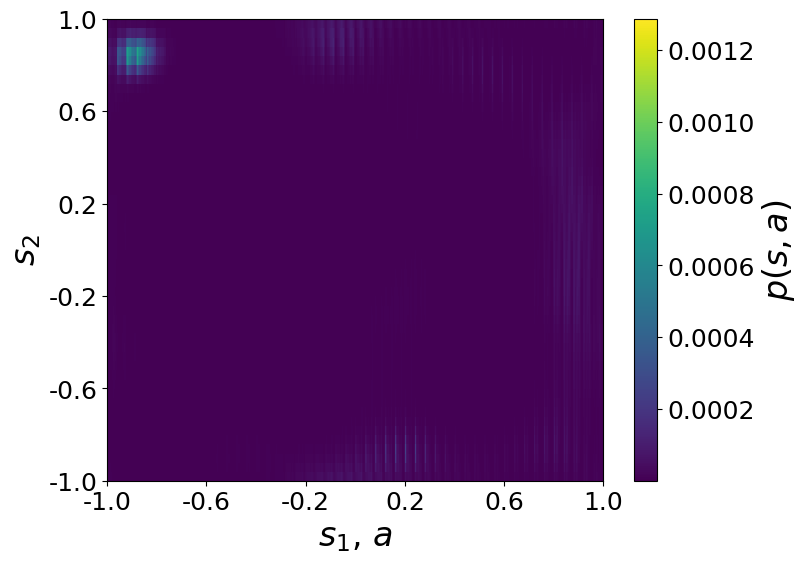}
        \caption{$t=25$}
    \end{subfigure}
    \hfill
    \begin{subfigure}[t]{0.38\textwidth}
        \centering
        \includegraphics[height=3.6cm]{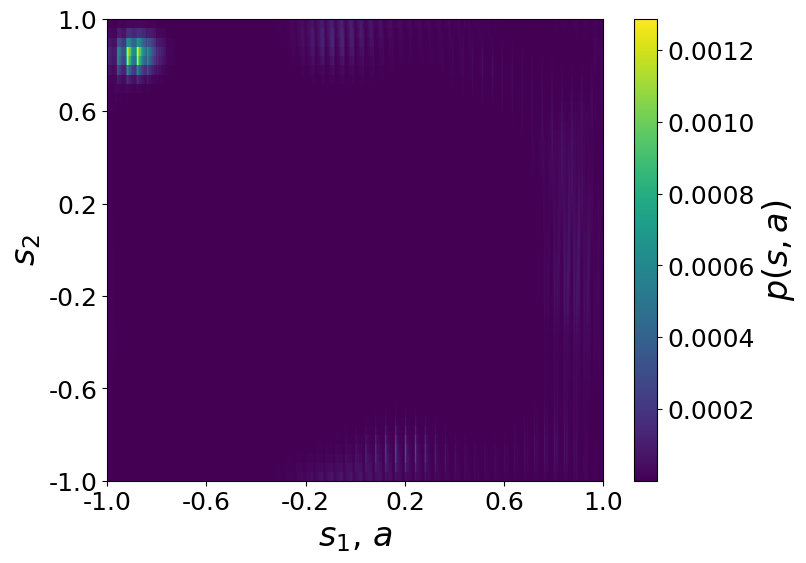}
        \caption{$t=50$}
    \end{subfigure}

    \begin{subfigure}[t]{0.3\textwidth}
        \centering
        \includegraphics[height=3.6cm, trim={0 0 4.5cm 0}, clip]{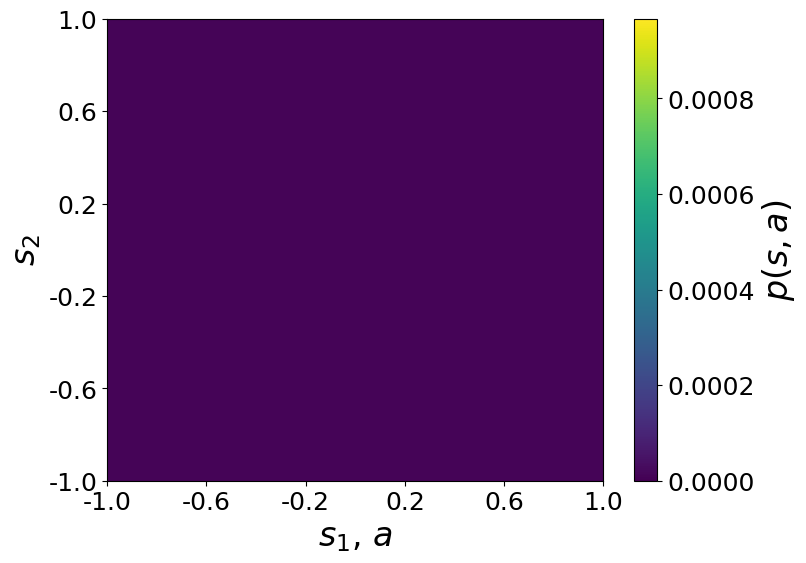}
        \caption{$t=0$}
    \end{subfigure}
    \hfill
    \begin{subfigure}[t]{0.3\textwidth}
        \centering
        \includegraphics[height=3.6cm, trim={0 0 4.5cm 0}, clip]{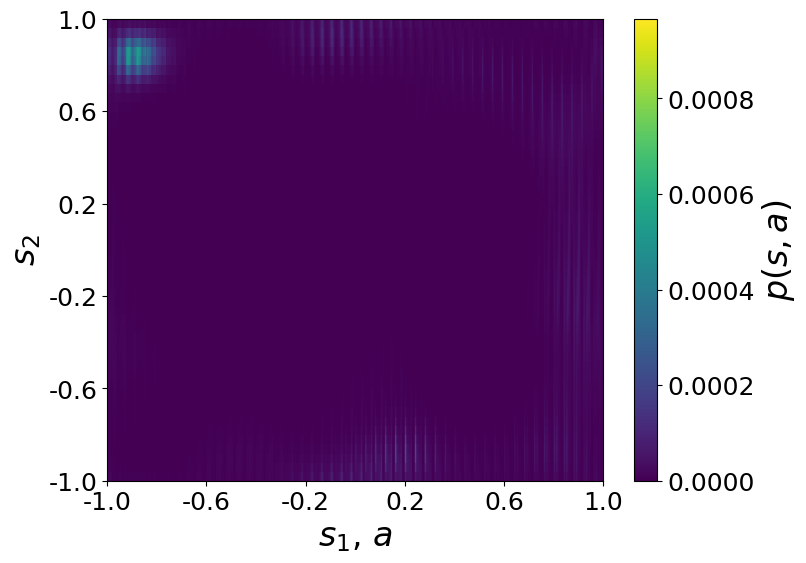}
        \caption{$t=25$}
    \end{subfigure}
    \hfill
    \begin{subfigure}[t]{0.38\textwidth}
        \centering
        \includegraphics[height=3.6cm]{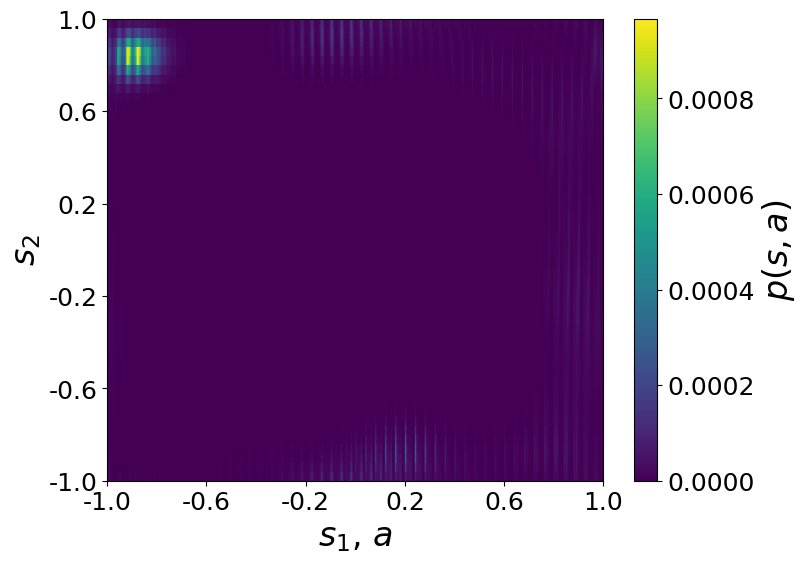}
        \caption{$t=50$}
    \end{subfigure}
    \caption{Mountain Car case. Joint probability density $p_t(\bss, \bsa)$ of state-action pairs over time, computed on a mesh with partition size $N=50$. The top panels (a)--(c) depict the ground-truth probability densities at selected time steps, while the bottom panels (d)--(f) show the corresponding inferred probability densities obtained using FP-IRL. Each panel represents the three-dimensional state-action space using primary grid indexed by the state variables $(s_1,s_2)$, with embedded sub-grids capturing variations over action variable $a$. Color intensity indicates the density magnitude over $\CalS \times \CalA$.}
    \label{fig:mountaincar_psa}
\end{figure}

\paragraph{FP-IRL results on modified problem}
We apply the full FP-IRL pipeline (\cref{alg:fpirl}) to this modified system. 
The potential function $\psi$ is inferred using VSI (\cref{sec:vsi}), with the inverse temperature $\beta$ in \cref{eq:fokker_planck_eq} treated as a fixed input. 
Using the inferred potential, we compute the  transition function, Q-function, and reward function are obtained via \cref{eq:mdp_transition,eq:conjecture_equivalence,eq:inverse_bellman}, respectively.

\begin{figure}[htbp]
    \centering
    \begin{subfigure}{0.49\textwidth}
        \centering
        \includegraphics[height=4.5cm]{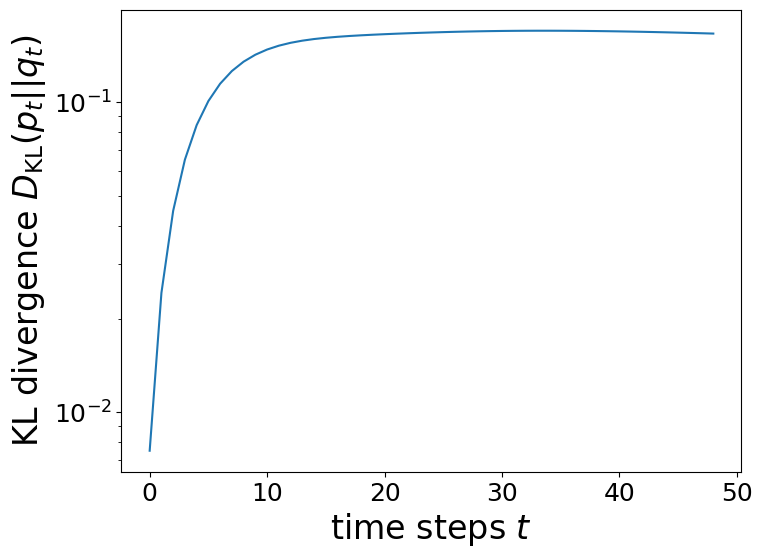}
        \caption{Full time interval ($t \in [0, 50]$).}
    \end{subfigure}
    \hfill
    \begin{subfigure}{0.49\textwidth}
        \centering
        \includegraphics[height=4.5cm]{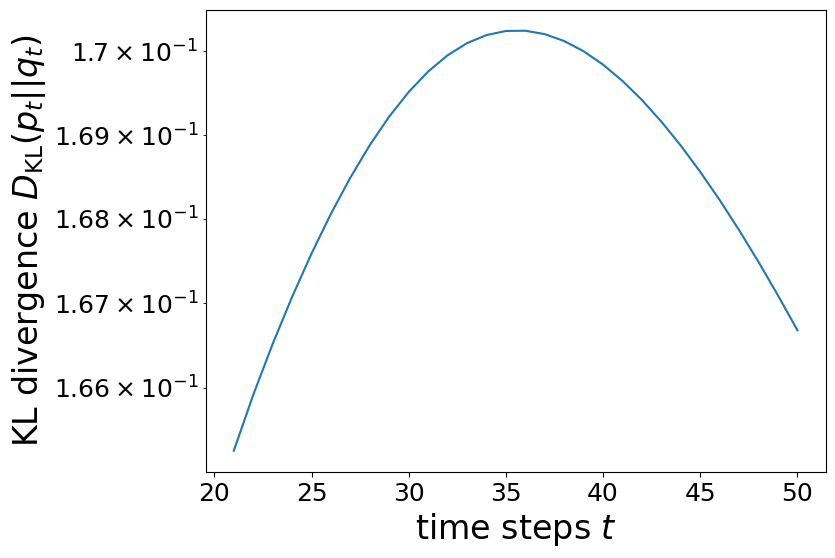}
        \caption{Zoomed-in interval ($t \in [20, 50]$).}
    \end{subfigure}
    \caption{Mountain Car case. KL divergence $\DKL(p_t^{\text{data}} || q_t)$ between the reference probability distribution $p_t^{\text{data}}$ from the data and the simulated distribution $q_t$, generated using the inferred policy and transition function. The divergence quantifies how closely the inferred dynamics match the observed data over time, with increasing divergence potentially reflecting accumulated inference errors.}
    \label{fig:mountaincar_KL}
\end{figure}

\begin{figure}[htbp]
    \centering
    \begin{subfigure}[t]{0.49\textwidth}
        \centering
        \includegraphics[height=4.5cm]{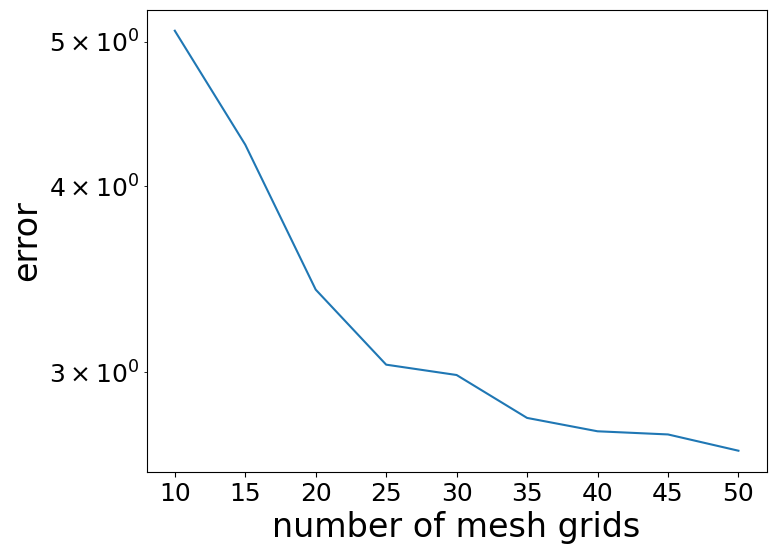}
        \caption{Error of potential function}
        \label{fig:mountaincar_psi_error}
    \end{subfigure}
    \hfill
    \begin{subfigure}[t]{0.49\textwidth}
        \centering
        \includegraphics[height=4.5cm]{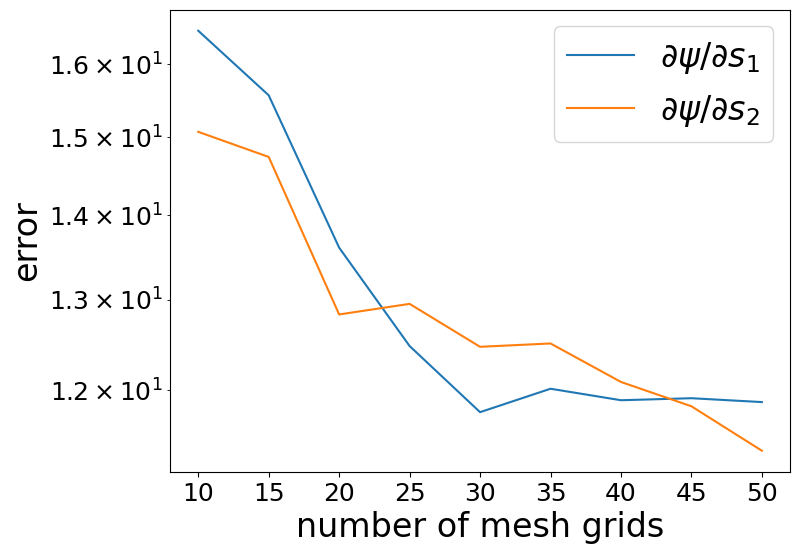}
        \caption{Error of the derivatives of potential function}
        \label{fig:mountaincar_dpsi_error}
    \end{subfigure}
    \caption{Mountain Car case. Convergence analysis of the inferred value function and its derivatives with respect to mesh resolution. The plots show how the errors in the estimated value function $\hat{Q}(\bss, \bsa) = -\hat{\psi}(\bss, \bsa)$ and its spatial derivatives $\partial \hat{Q} / \partial \bss$ decrease as the number of partitions $N$ in each dimension increases. This demonstrates the expected convergence behavior of the FP-IRL framework under mesh refinement.}
    \label{fig:mountaincar_convergence}
\end{figure}

FP-IRL results are shown in \cref{fig:moutaincar_q_value_vsi,fig:moutaincar_reward_vsi,fig:moutaincar_policy_vsi}, obtained using the highest-resolution mesh with $N=50$. These are compared against the original ground-truth functions in \cref{fig:moutaincar_q_value_gt,fig:mountain_car_reward_gt,fig:mountain_car_policy_gt}, demonstrating excellent agreement across all quantities.
\Cref{fig:mountaincar_psa} further compares the joint probability densities $p_t(\bss, \bsa)$ between the inferred and ground-truth systems, showing consistent matching over time. 
The KL divergence $D_{\text{KL}}(p_t || q_t )$ between the observed and simulated densities is plotted in \cref{fig:mountaincar_KL}. As expected, divergence increases with time due to the cumulative effect of inference errors and decreases slowly as $t > 35$, similar to the behavior observed in the synthetic Grid World example.
Finally, \cref{fig:mountaincar_convergence} presents a convergence study with respect to mesh resolution. Both the error in the inferred potential function and its derivatives decrease as the mesh is refined, confirming a clear trend toward convergence.

\FloatBarrier
\section{Discussion}
\label{sec:discussion}

\subsection{Significance}
\label{sec:discussion_significance}

We have introduced a novel FP-IRL framework by introducing a correspondence between FP physics and a class of physics-based MDPs.
This connection enables the development of a physics-constrained IRL algorithm that infers both the reward function and transition dynamics from observed behavior, without requiring explicit trajectory simulation. 
This fusion of physics and IRL offers several key advantages. 

\paragraph{Interpretability through physics} 
The incorporation of FP dynamics allows us to ground the inferred value function in physical principles. By inferring the FP PDE via VSI and invoking \cref{thm:equivalence}, we obtain a potential (or value) function whose gradient governs drift, encapsulating the systematic tendencies of agent behavior. This physical interpretability enables the decomposition of learned dynamics into meaningful components such as:
drift representing systematic directional bias and diffusion quantifying randomness or stochasticity in the dynamics.

\paragraph{Addressing ill-posedness in IRL}
Traditional IRL suffers from severe ill-posedness, where many reward-transition pairs may explain the same observed behavior. 
This issue is exacerbated when the transition dynamics are unknown and must be empirically estimated, often leading to poor generalization in sparsely observed regions of the state-action space.
By incorporating FP dynamics as a physics-based constraint, we narrow the solution space to only those reward-transition pairs consistent with underlying physical laws. This constraint reduces ambiguity and improves practical identifiability within the assumed FP model class, and enhances robustness especially in scientific settings where FP dynamics are known or hypothesized to govern the system.

\paragraph{Improved computational efficiency}
Standard IRL methods rely on nested optimization loops: an outer loop updating the reward function and an inner loop solving forward RL problems to optimize the policy (see \cref{alg:generic_IRL}). This two-tier structure is computationally expensive, particularly when deep RL algorithms are used in the inner loop.
In contrast, FP-IRL replaces this costly optimization with a regression-based inversion problem, leveraging the known structure of FP dynamics. Since this regression uses basis expansions (e.g., Hermite polynomials) and avoids expensive policy iteration, it is generally faster to solve, especially when the basis size is modest. It is also more stable, avoiding the convergence issues of adversarial optimization, and less data-hungry, due to structural regularization from physics.
The computational complexity comparison is summarized in \cref{tab:comutational_complexity}.

\begin{table}[htb] 
\centering
\caption{Comparative analysis of computational complexity for FP-IRL versus standard IRL methods. 
Let $n_d \approx m\tau$ denote the number of data points, where $m$ is the number of trajectories and $\tau$ is the trajectory length. 
Let $n_e = \prod_{i=1}^{d} n_{e, i}$ represent the number of finite elements in the VSI mesh, and $n_b = \prod_{i=1}^{d} n_{e, i} n_{h, i}$ denote the number of basis functions used. 
For the tabular RL method, $|\CalS|$ and $|\CalA|$ are the sizes of the (discretized) state and action spaces, respectively.
When using uniform grid discretization, $|\CalS| = \prod_{i=1}^{d_s} n_{e, i}$ and $|\CalA| \approx \prod_{i=1}^{d_a} n_{e, i}$.
For standard IRL methods based on neural networks, let $n_p = n_l n_n^2$ denote the number of parameters in a network with $n_l$ layers and $n_n$ neurons per layer (typically $n_p \gg n_b$). 
The number of training epochs is denoted by $k$. 
The table summarizes dominant cost terms in each stage of computation, 
and $\CalO(\mathrm{LR})$ and $\CalO(\mathrm{RL})$ denote the computational costs of the linear regression step in VSI and the RL step in IRL, respectively.
}
\label{tab:comutational_complexity}
\resizebox{\textwidth}{!}{%
\setlength\extrarowheight{2.5pt}
\renewcommand{\arraystretch}{1.1}
\begin{tabular}{lllc|lllc}
\hline
\multicolumn{4}{c|}{FP-IRL}                                                                                                                                                                                                                       & \multicolumn{4}{c}{Standard IRL}                                                                                                                                                                                                         \\ \hline
\multicolumn{1}{l|}{\multirow{4}{*}{VSI}} & \multicolumn{2}{l|}{Binning}                                                                                                                                        & $\CalO(n_d n_e)$                & \multicolumn{3}{l|}{Transition modeling}                                                                                                                                                                          & $\CalO(k n_d n_p)$   \\ \cline{5-8} 
\multicolumn{1}{l|}{}                     & \multicolumn{2}{l|}{Basis generation}                                                                                                                               & $\CalO(n_b n_e \tau)$           & \multicolumn{1}{l|}{\multirow{5}{*}{\begin{tabular}[c]{@{}l@{}}IRL loop\\ $k_{\text{IRL}} \times \cdots$\end{tabular}}} & \multicolumn{1}{l|}{\multirow{2}{*}{RL}}         & \multicolumn{1}{l|}{Tabular method}         & $\CalO(k |\CalS|^2 |\CalA|)$ \\ \cline{2-4}
\multicolumn{1}{l|}{}                     & \multicolumn{1}{l|}{\multirow{2}{*}{\begin{tabular}[c]{@{}l@{}}FP PDE inference by \\ linear regression (LR)\end{tabular}}} & \multicolumn{1}{l|}{Matrix method}    & $\CalO(\tau n_e n_b^2 + n_b^3)$ & \multicolumn{1}{l|}{}                                                                                            & \multicolumn{1}{l|}{}                            & \multicolumn{1}{l|}{Policy gradient}        & $\CalO(k n_p)$       \\ \cline{6-8} 
\multicolumn{1}{l|}{}                     & \multicolumn{1}{l|}{}                                                                                                       & \multicolumn{1}{l|}{Gradient descent} & $\CalO(k_{LR} n_b)$             & \multicolumn{1}{l|}{}                                                                                            & \multicolumn{2}{l|}{Simulation}                                                                & $\CalO( n n_p)$      \\ \cline{1-4}
\multicolumn{1}{l|}{\multirow{2}{*}{IRL}} & \multicolumn{2}{l|}{Reward inference}                                                                                                                               & $\CalO( n_e^2)$                 & \multicolumn{1}{l|}{}                                                                                            & \multicolumn{2}{l|}{\begin{tabular}[c]{@{}l@{}}Comparison \\ (Occupancy measure)\end{tabular}} & $\CalO(n_d)$         \\
\multicolumn{1}{l|}{}                     & \multicolumn{2}{l|}{Policy inference}                                                                                                                               & $\CalO( n_e)$                   & \multicolumn{1}{l|}{}                                                                                            & \multicolumn{2}{l|}{Optimization}                                                              & $\CalO(n_p)$         \\ \hline
\multicolumn{1}{l|}{Dominant}             & \multicolumn{3}{c|}{$\CalO(\mathrm{LR})$}                                                                                                                                                                      & \multicolumn{1}{l|}{Dominant}                                                                                    & \multicolumn{3}{c}{$k_{\text{IRL}} \CalO(\mathrm{RL})$}                                                                        \\ \hline
\end{tabular}%
}
\end{table}

\begin{algorithm}[htbp]
    \caption{Standard IRL Algorithm}
    \label{alg:generic_IRL}
    \KwIn{A Markov decision process without reward functions $\CalM \setminus \{ R\}$, observed trajectories $\CalD$.}
    \KwOut{Estimated reward function $R$ and corresponding policy $\pi$.}
    \If{transition dynamics are unknown}{
        Estimate the transition function using $\mathcal{D}$\;
    }
    Initialize reward function $R$\;
    \While{reward function has not converged}{
        Apply an RL algorithm to solve MDP given current $R$\; \label{alg_line:RL}
        Generate trajectories using the policy from Line 6\;
        Update $R$ by minimizing a predefined discrepancy measure between the learned and observed trajectories\;
    }
\end{algorithm}

\paragraph{Broader applicability}
FP-IRL is particularly well-suited to domains where transitions are not explicitly known but are governed by FP-like stochastic processes. In biology, for instance, the migration of cancer cells, immune cells, or bacteria often adheres to FP-type dynamics~\cite{bressloff2014stochastic}. Consequently, FP-IRL enables reward inference for cell agents without needing explicit models of cell motion, mechanistic interpretation of inferred behavior, and generalization beyond observed trajectories.
Beyond biology, FP dynamics also arise in Brownian motion~\cite{Keilson1952OnBM}, collective swarming~\cite{Correll2015Swarming} and crowd dynamics~\cite{Dogbe2010Crowd}, and pattern formation or morphogenesis~\cite{Garikipati2017Perspectives}. In these domains, FP-IRL provides a promising path toward interpretable and physically grounded agent-based modeling.

\subsection{Limitations}
\label{sec:limitations}
While FP-IRL offers a novel and interpretable framework for IRL grounded in physics, several limitations remain.

\paragraph{Dependence on FP dynamics}
FP-IRL fundamentally assumes that the system dynamics adhere to the FP formulation governed by free energy principles. This assumption often requires prior domain knowledge or empirical justification and limits the method's applicability to systems with well-characterized continuous stochastic dynamics.
Furthermore, because Brownian dynamics are typically posed in unbounded domains, the framework assumes an open unbounded domain $\mathcal{S}\times\mathcal{A} \subset\mathbb{R}^n$ for the state-action space. 
In practice, we impose periodic boundary conditions to approximate this behavior, but extending the method to more realistic boundary conditions, such as reflecting or absorbing walls, would require incorporating more complex stochastic processes (e.g., reflected Brownian motion).
FP-IRL assumes a locally Gaussian transition model, derived from a small time-step first-order approximation of the underlying stochastic dynamics. This structure is consistent with the FP transition probability density function for linear potentials. The Orenstein--Uhlenbeck process with quadratic potentials also leads to a transition function with underlying Gaussian structure \cite{risken1996fokker}.
Systems with heavy-tailed increments, jumps, switching modes, or other non-diffusive effects generally require replacing the FP equation with a more appropriate forward model,
which is outside the scope of the present work. 
Finally, our formulation emphasizes conservative (gradient) drift fields, which are compatible with equilibrium/detailed-balance structure; driven non-equilibrium steady states with non-conservative drift and circulating probability currents are therefore not covered here, and handling such settings would require extending the framework to infer general drift fields. %

\paragraph{Assumption of continuity}
FP-IRL operates within the PDE framework and therefore assumes that both state and action variables are continuous. This makes it unsuitable for problems defined over discrete or coarsely quantized state-action spaces, where accurate estimation of the potential function and its derivatives becomes infeasible. As demonstrated in our convergence analysis, fine discretization is critical for reliable recovery, but this increases the computational burden significantly.

\paragraph{Scalability and curse of dimensionality}
Although FP-IRL extends conceptually to high-dimensional state-action spaces, its practical implementation---particularly the VSI procedure---relies on finite element methods, which scale poorly with dimensionality. The number of mesh elements and basis functions grows exponentially with the dimension of state-action space $d$, leading to significant computational overhead. One potential remedy is to replace finite element basis functions with neural network surrogates for the value function, which may offer better scalability while retaining structure from physics.

\paragraph{Limitation of single-agent modeling}
The current formulation of FP-IRL is built on single-agent dynamics and assumes independent agents. As such, it cannot model systems with explicit inter-agent interactions, such as swarms, coordinated groups, or game-theoretic settings with strategic behavior. Extending FP-IRL to multi-agent systems with interactions remains an open direction for future work.

\section{Conclusions}
\label{sec:conclusion}

We have presented FP-IRL, a novel physics-constrained IRL framework that bridges principles from stochastic physics and RL. By conjecturing an equivalence between the FP equation and the MDP, FP-IRL enables the inference of both the reward and transition functions from trajectory data, without requiring direct access to the environment's dynamics or iterative policy optimization.

Our approach brings three key advantages:
\begin{enumerate}
\item it removes the dependency on sampled transitions or black-box simulators;
\item it retains interpretability through physically meaningful quantities such as drift and diffusion; and
\item it offers computational efficiency by transforming IRL into a regression problem solved via VSI.
\end{enumerate}

We validated FP-IRL on both a synthetic Grid World and a modified version of the Mountain Car benchmark adapted to FP dynamics. Across both settings, FP-IRL accurately recovers the underlying reward structure, transition dynamics, and optimal policy. We observed systematic convergence of the inferred quantities under mesh refinement, highlighting the method's robustness and consistency. Furthermore, KL divergence metrics and visual comparisons confirmed close agreement between observed and simulated behavior under the inferred policy. 

While the method currently assumes continuous FP dynamics, it opens promising directions for future work in high-dimensional systems, interacting agents, and neural surrogates for value function approximation. FP-IRL is particularly well-suited for applications in biology, physics, and complex decision-making systems where physical principles govern behavior but mechanistic knowledge is partial or incomplete. 

Overall, FP-IRL contributes a new class of physics-informed IRL algorithms that enhances both the interpretability and generalizability of learned agent behavior in scientific domains.

\section*{Acknowledgments}
This research is supported in part by the W. M. Keck Foundation. Additional support is provided through computational resources and services provided by Advanced Research Computing at the University of Michigan, Ann Arbor.

\FloatBarrier
\bibliography{reference.bib}
\bibliographystyle{elsarticle-num-names}

\newpage
\appendix

\renewcommand{\thesection}{\Alph{section}} %
\makeatletter
\def\@seccntformat#1{\@ifundefined{#1@cntformat}%
   {\csname the#1\endcsname.\hspace{0.5em}}%
   {\csname #1@cntformat\endcsname}}%
\newcommand\section@cntformat{\appendixname\ \thesection.\hspace{0.5em}}
\makeatother

\section{``Triangle Inequality-like'' Result for Wasserstein-2 Distance}
\label{appsec:wasserstein_chain_rule}
Let $\bsx, \bsy \in \Omega$ be random vectors distributed according to $p(\bsx)$ and $q(\bsy)$, respectively.
The Wasserstein-2 distance between $p$ and $q$ is defined as:
\begin{align}
    W_2^2(p, q) = \inf_{\gamma \in \Gamma(p, q)} \EE_{(\bsx, \bsy) \sim \gamma} \left[ \| \bsx - \bsy \|_2^2 \right],
\end{align}
where $\Gamma(p, q)$ denotes the set of all couplings (joint distributions) $\gamma(\bsx, \bsy)$ on $\Omega \times \Omega$ with marginals $p(\bsx)$ and $q(\bsy)$. 
That is, $\gamma \in \Gamma$ satisfies both 
$\int_{\Omega} \gamma(\bsx, \bsy) \, \mathrm{d} \bsy = p(\bsx)$ and $\int_{\Omega} \gamma(\bsx, \bsy) \, \mathrm{d} \bsx = q(\bsy)$.

We first partition $\bsx$ and $\bsy$ as $\bsx = (\bss_x, \bsa_x)$ and $\bsy = (\bss_y, \bsa_y)$.
Let $p_{\bss}$ and $q_{\bss}$ denote the marginal distributions over states $\bss$ for $p$ and $q$, respectively, and let $p_{\bsa | \bss}$ and $q_{\bsa | \bss}$ denote their corresponding conditional distributions over actions given states. 
A ``triangle inequality-like'' result for the Wasserstein-2 distance states:
\begin{align}
    W_2^2(p, q) \leq W_2^2(p_{\bss}, q_{\bss}) + \EE_{(\bss_x, \bss_y) \sim \gamma_s^{\ast}} \left[ W_2^2(p_{\bsa | \bss_x}, q_{\bsa | \bss_y}) \right],
\end{align}
where $\gamma_s^{\ast}$ is the optimal coupling between the state marginals $p_{\bss}$ and $q_{\bss}$ in $W_2^2(p_{\bss}, q_{\bss})$.

\begin{proof}
Let 
$\mathcal{S} \subseteq \mathbb{R}^{d_s}$ and $\mathcal{A} \subseteq \mathbb{R}^{d_a}$ be Borel sets, and define $\Omega = \mathcal{S} \times \mathcal{A}$.
Suppose $p, q \in \mathcal{P}_2(\Omega)$ are absolutely continuous with respect to Lebesgue measure on $\mathbb{R}^{d}$ where $d = d_s + d_a$. 
In particular, both admit densities $p(\bss, \bsa)$ and $q(\bss, \bsa)$ that factor as:
\begin{align*}
    p(\bss,\bsa) = p_{\bss}(\bss)\, p_{\bsa|\bss}(\bsa|\bss), 
    \qquad
    q(\bss,\bsa) = q_{\bss}(\bss)\, q_{\bsa|\bss}(\bsa|\bss).
\end{align*}
Let $\Gamma(p, q)$ denote the set of couplings of $p$ and $q$ on $\Omega \times \Omega$, and $\Gamma_s(p_{\bss}, q_{\bss})$ the set of couplings of the state marginals.

According to \citet[Theorem 3.7]{gangbo1996geometry}, the unique existence of an optimal transport map for the Wasserstein-2 metric follows from the use of a quadratic cost function in its definition.

For any $\gamma \in \Gamma(p, q)$, let $\gamma_s$ denote its marginal on $(\bss_x, \bss_y)$.
By disintegration, there exists a conditional law $\eta(\bsa_x, \bsa_y | \bss_x, \bss_y)$ such that:
\begin{align*}
    \gamma(\bss_x, \bsa_x, \bss_y, \bsa_y) = \gamma_s(\bss_x, \bss_y) \, \eta(\bsa_x, \bsa_y | \bss_x, \bss_y),
\end{align*}
for almost every $(\bss_x, \bss_y)$ and conditional $\eta(\cdot\mid \bss_x, \bss_y) \in \Gamma(p_{\bsa|\bss_x}, q_{\bsa|\bss_y})$.

Since the quadratic cost separates additively,
\begin{align*}
    \| (\bss_x, \bsa_x) - (\bss_y, \bsa_y) \|_2^2 = \| \bss_x - \bss_y \|_2^2 + \| \bsa_x - \bsa_y \|_2^2,
\end{align*}
the expected transport cost under $\gamma$ decomposes as:
\begin{align*}
    \EE_{\gamma}\!\left[\|(\bss_x,\bsa_x)-(\bss_y,\bsa_y)\|_2^2\right]
    &= \EE_{(\bss_x,\bss_y)\sim\gamma_s}\!\big[\|\bss_x-\bss_y\|_2^2\big] \\
    &\quad+ \EE_{(\bss_x,\bss_y)\sim\gamma_s}
    \left[ \EE_{(\bsa_x,\bsa_y)\sim\eta(\cdot|\bss_x,\bss_y)}
    \left[ \| \bsa_x-\bsa_y \|_2^2 \right] \right].
\end{align*}

Let $\gamma_s^\ast \in \Gamma_s(p_{\bss},q_{\bss})$ be an optimal coupling of the state marginals, achieving
\begin{align*}
    W_2^2(p_{\bss},q_{\bss})
    = \EE_{(\bss_x,\bss_y)\sim \gamma_s^\ast}\!\big[\|\bss_x-\bss_y\|_2^2\big].
\end{align*}

Because $p,q$ are densities of absolutely continuous measure on $\Omega$, each conditional $p_{\bsa|\bss}$ is absolutely continuous on $\mathcal{A}\subseteq\mathbb{R}^{d_a}$.
Thus, there exists a unique optimal transport between $p_{\bsa|\bss_x}$ and $q_{\bsa|\bss_y}$. 

Let $\eta^\ast(\cdot|\bss_x,\bss_y)$ denote the corresponding optimal coupling, which satisfies
\begin{align*}
W_2^2(p_{\bsa|\bss_x},q_{\bsa|\bss_y})
= \EE_{(\bsa_x,\bsa_y)\sim\eta^\ast(\cdot|\bss_x,\bss_y)}\!\big[\|\bsa_x-\bsa_y\|_2^2\big].
\end{align*}

Define the joint plan
\begin{align*}
    \widetilde{\gamma}(\bss_x,\bsa_x,\bss_y,\bsa_y)
    := \gamma_s^\ast(\bss_x,\bss_y)\, \eta^\ast(\bsa_x,\bsa_y|\bss_x,\bss_y),
\end{align*}
where by construction $\widetilde{\gamma}\in\Gamma(p,q)$. The cost of $\widetilde{\gamma}$ is
\begin{align*}
    \EE_{\widetilde{\gamma}}\!\left[\|(\bss_x,\bsa_x)-(\bss_y,\bsa_y)\|_2^2\right]
    &= W_2^2(p_{\bss},q_{\bss}) 
    + \EE_{(\bss_x,\bss_y)\sim \gamma_s^\ast}
    \left[ W_2^2(p_{\bsa|\bss_x},q_{\bsa|\bss_y}) \right].
\end{align*}
Since $W_2^2(p,q)$ is the minimum transport cost over all $\gamma\in\Gamma(p,q)$, we conclude
\begin{align*}
    W_2^2(p,q) \leq W_2^2(p_{\bss},q_{\bss})
    + \EE_{(\bss_x,\bss_y)\sim \gamma_s^\ast}
    \left[ W_2^2(p_{\bsa|\bss_x},q_{\bsa|\bss_y}) \right].
\end{align*}
\end{proof}

\section{Inverse Bellman Operator}
\label{appsec:inverse_bellman}
We provide a proof for \cref{thm:inverse_bellman}.
Our approach follows a similar structure to the proof of Lemma 3.1 in Appendix 2 of \citet{garg2021iqlearn}, though there is a distinction in the definition of the inverse Bellman operator. 

In \citet{garg2021iqlearn}, the
inverse \textit{soft} Bellman operator is defined using a soft Q-function: 
\begin{align}
    R(\bss, \bsa) = (\CalT^{\pi}_{\text{soft}} Q^{\pi})(\bss, \bsa) = Q^{\pi}_{\text{soft}}(\bss, \bsa) - \gamma \EE_{\subalign{\bss' \sim& T(\cdot | \bss, \bsa) \\ \bsa' \sim& \pi(\cdot | \bss')}} \[ Q^{\pi}_{\text{soft}}(\bss', \bsa') - \log \pi(a' | s')\], 
\end{align}
where $Q^{\pi}_{\text{soft}}$ satisfies the soft Bellman equation and includes entropy regularization. In contrast, our formulation uses the conventional Bellman expectation and defines the operator as:
\begin{align}
    R(\bss, \bsa) = (\CalT^{\pi} Q^{\pi})(\bss, \bsa) = Q^{\pi}(\bss, \bsa) - \gamma \EE_{\subalign{\bss' \sim& T(\cdot | \bss, \bsa) \\ \bsa' \sim& \pi(\cdot | \bss')}} \[ Q^{\pi}(\bss', \bsa') \],
\end{align}
where $Q^{\pi}$ denotes the standard action-value function for policy $\pi$, without entropy terms.

\begin{lemma}
\label{lem:invertible}
    Let $\bsA$ be a square matrix such that $ || \bsA ||< 1$ for some consistent matrix norm. Then $\bsI-\bsA$ is nonsingular (i.e., invertible).
\end{lemma}
\begin{proof}
    We prove by contradiction. Suppose that $\bsI-\bsA$ is singular. Then there exists a nonzero vector $\bsx \neq 0$ such that $(\bsI-\bsA) \bsx = 0$, $\implies \bsx=\bsA\bsx$. Taking norms on both sides yields $||\bsx|| = ||\bsA \bsx || \leq ||\bsA|| \ ||\bsx||$. Dividing both sides by $||\bsx||\geq 0$, we obtain $1\leq || \bsA ||$, which contradicts the assumption that $||\bsA|| < 1$. Therefore, $\bsI-\bsA$ must be nonsingular.
\end{proof}

\inversebellman*

\begin{proof}
    For a given $T(\bss' | \bss, \bsa)$ and $\pi(\bsa | \bss)$, the joint transition function is $T_{\text{MP}}(\bss', \bsa'| \bss, \bsa) = T(\bss' | \bss, \bsa) \pi(\bsa' | \bss')$. In the discrete form, the inverse Bellman operator can be written in matrix form:
    \begin{align}
        \bsr = \bsq -  \gamma \bsT_{\text{MP}} \bsq = (\bsI - \gamma \bsT_{\text{MP}}) \bsq,
    \end{align}
    where $\bsr \in \RR^{n_s \cdot n_a}$ is the reward vector, $\bsq \in \RR^{n_s \cdot n_a}$ is the flattened state-action value vector, $\bsT_{\text{MP}} \in \RR^{(n_s \cdot n_a) \times (n_s \cdot n_a)}$ is the joint transition matrix, and
    $n_s = |\CalS|$ and $n_a = |\CalA|$ are the number of discretized states and actions, respectively. 
    By construction, $\bsT_{\text{MP}}$ is a stochastic matrix, where its rows are probability distributions and so  $|| \bsT_{\text{MP}} ||_1 = 1$. Since $\gamma \in [0, 1)$, we have $|| \gamma \bsT_{\text{MP}} ||_1 < 1$.
    Therefore, by \cref{lem:invertible}, $\bsI - \gamma \bsT_{\text{MP}}$ is nonsingular.
    It follows that $\bsq=(\bsI - \gamma \bsT_{\text{MP}})^{-1}\bsr$ has a unique solution for any $\bsr$, and vice versa. Hence, the inverse Bellman operator $\CalT$ is bijective under fixed $T$ and $\pi$.
\end{proof}

\end{document}